\pdfoutput=1

\documentclass[11pt]{article}

\usepackage{EMNLP2023}

\usepackage{times}
\usepackage{latexsym}

\usepackage[T1]{fontenc}

\usepackage[utf8]{inputenc}

\usepackage{microtype}

\usepackage{inconsolata}

\usepackage{amsmath,amsfonts,bm}
\usepackage{booktabs} %
\usepackage{url}
\usepackage{graphicx}
\usepackage{array}
\usepackage{color}
\usepackage{makecell}
\usepackage{multirow}
\usepackage{xspace}
\usepackage{colortbl}
\usepackage{subcaption}
\usepackage{algpseudocode}
\usepackage[noend,ruled]{algorithm2e}
\usepackage{setspace}
\usepackage{varwidth}

\usepackage{soul}
\usepackage{pifont}
\usepackage{graphics}

\newcommand{\ours}{CMTrans\xspace}
\newcommand{\CC}{Comp-Corp\xspace}
\newcommand{\hlours}[1]{%
  \begingroup
  \sethlcolor{#1}%
  \hl{ CMTrans }%
  \endgroup
}

\newcommand{\edit}[1]{\textcolor{black}{#1}}

\newcommand{\start}[1]{\vspace{.3mm}\noindent{{\bf #1}.}}

\newcommand{\downv}{\vspace{-.0cm}}
\newcommand{\upv}{\vspace{-.0cm}}

\definecolor{gred}{RGB}{255,102,102}
\definecolor{gblue}{RGB}{51,102,255}
\definecolor{gyellow}{RGB}{244,180,0}
\definecolor{ggreen}{RGB}{15,157,88}
\definecolor{ggrey}{RGB}{115,115,115}
\definecolor{na}{gray}{0.9}
\definecolor{LightYellow}{RGB}{255,255,191}
\definecolor{OrangeRed}{rgb}{1.0, 0.27, 0.0}
\definecolor{midnightgreen}{rgb}{0.0, 0.29, 0.33}
\definecolor{darkgreen}{rgb}{0.0, 0.42, 0.24}
\definecolor{editblue}{rgb}{0.27734375, 0.2734375, 0.6484375}

\definecolor{diagramRed}{RGB}{246,193,193}
\definecolor{diagramPurple}{RGB}{224,224,253}
\definecolor{diagramOrange}{RGB}{244,222,176}
\definecolor{diagramGrey}{RGB}{236,236,236}

\newcommand{\ctext}[2]{%
  \begingroup
  \sethlcolor{#1}%
  \hl{ #2 }%
  \endgroup
}

\usepackage{amsmath,amsfonts,bm}

\def\eqref#1{equation~\ref{#1}}

\def\1{\bm{1}}

\DeclareMathAlphabet{\mathsfit}{\encodingdefault}{\sfdefault}{m}{sl}
\SetMathAlphabet{\mathsfit}{bold}{\encodingdefault}{\sfdefault}{bx}{n}

\title{Data Augmentation for Code Translation with \\Comparable Corpora and Multiple References}

\author{Yiqing Xie ~~ Atharva Naik ~~ Daniel Fried ~~ Carolyn Ros\'e \\
  Language Technologies Institute \\
  Carnegie Mellon University \\
  \texttt{\{yiqingxi, arnaik, dfried, cprose\}@cs.cmu.edu}}

\begin{document}
\maketitle

\begin{abstract}
One major challenge of translating code between programming languages is that parallel training data is often limited.
To overcome this challenge, we present two data augmentation techniques, one that builds comparable corpora (i.e., code pairs with similar functionality), and another that augments existing parallel data with multiple reference translations.
\edit{Specifically, we build and analyze multiple types of comparable corpora, including programs generated from natural language documentation using a code generation model.}
Furthermore, to reduce overfitting to a single reference translation, we automatically generate additional translation references for available parallel data and filter the translations by unit tests, which increases variation in target translations.
Experiments show that our data augmentation techniques significantly improve CodeT5 for translation between Java, Python, and C++ by an average of 7.5\% Computational Accuracy (CA@1), which verifies the correctness of translations by execution.\footnote{Code available at https://github.com/Veronicium/CMTrans.}
\end{abstract}

\section{Introduction}

\begin{figure}[!tp]
    \centering
    \vspace{-0.1cm}
    \includegraphics[width=0.95\linewidth]{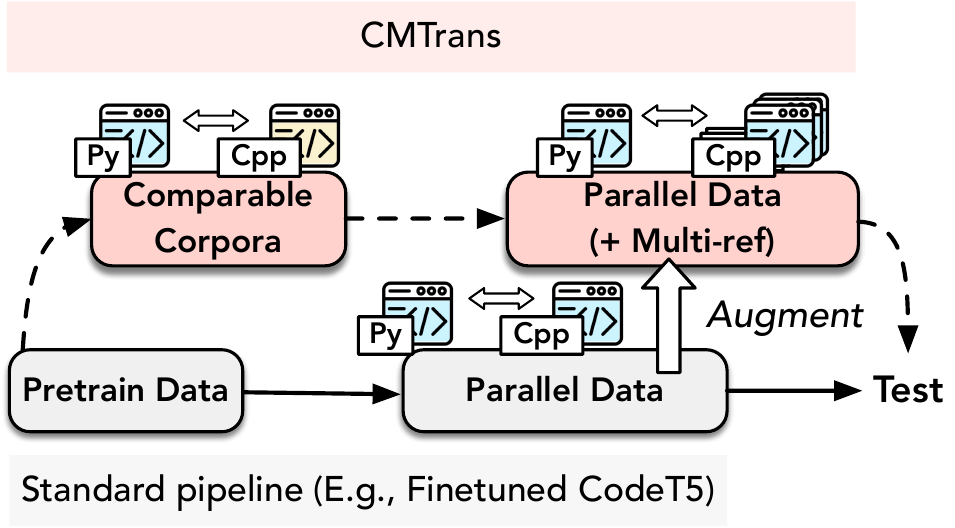}
    \upv
    \vspace{-0.1cm}
    \caption{The \ctext{diagramGrey}{standard pipeline} for code translation and the pipeline of \hlours{diagramRed}. The comparable corpora are both naturally occurring and model generated. We generate multiple references by our method.
    }
    \label{diagram_intro}
    \downv
    \vspace{-0.3cm}
\end{figure}

\begin{figure*}[ht]
    \centering
    \includegraphics[width=\linewidth]{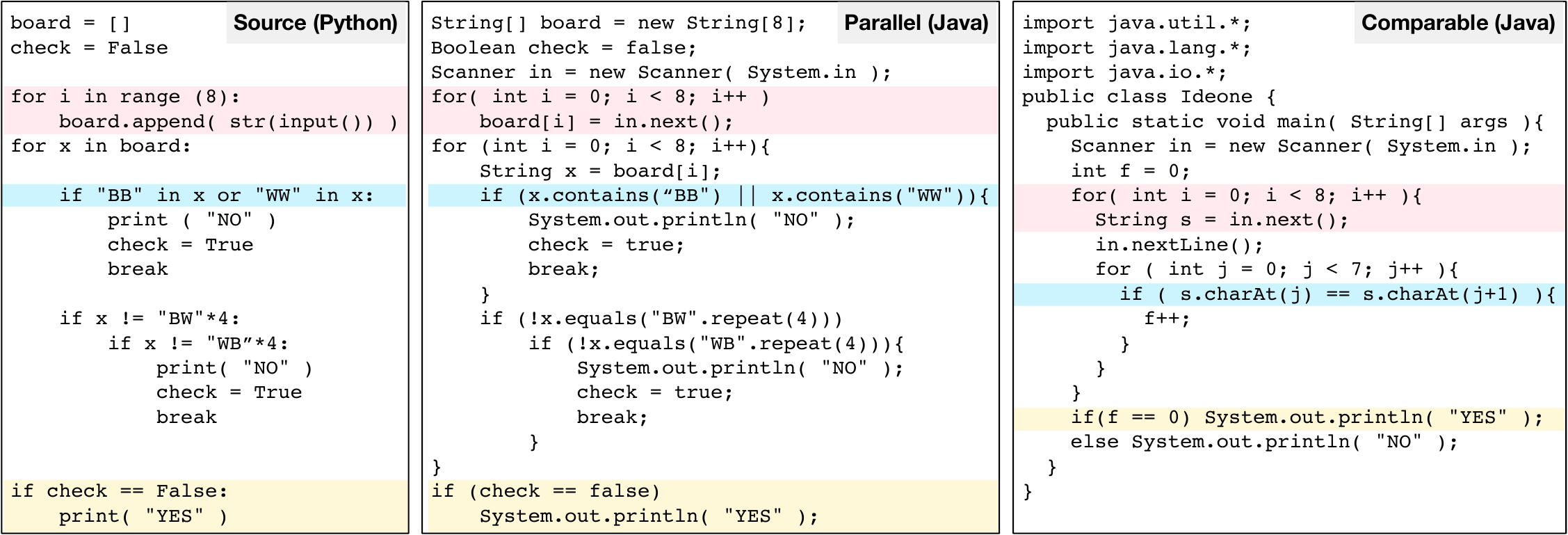}
    \vspace{-0.5em}
    \upv
    \caption{An example of parallel and comparable data. Parallel examples are line-by-line aligned. Programs in a comparable example may have different algorithms and structures (e.g., global code vs. class in this case), but may still contain lines that can be matched, as highlighted in pink, blue, and yellow.}
    \label{example_intro}
    \downv
\end{figure*}

Code translation is a special type of machine translation that translates between programming languages.
It is widely applied in software engineering to migrate a codebase into another programming language.
Recent code translation models typically follow the pretrain-finetune pipeline, as shown in \autoref{diagram_intro}.
In pretraining, with denoising objectives such as masked span or identifier prediction \cite{plbart,codet5,transcoder}, the model learns to produce sequences in both languages.
When finetuned on parallel data \cite{avatar}, which are program pairs in the source and target language that are aligned line-by-line, the model learns functional equivalence: identifying programs with the same functionality,
either in the same language or between languages.
We show an example of parallel data in \autoref{example_intro}.

One major challenge of code translation is that parallel data is typically limited. For instance, the TransCoder dataset \cite{transcoder} only contains 466 Python-C++ pairs. Constructing parallel data requires substantial human effort and cannot be easily scaled. 
With limited fine-tuning examples, it is difficult for a model to learn functional equivalence across programming styles and domains.

To mitigate the data scarcity issue, we hypothesize that the functional equivalence between languages can also be learned from \emph{comparable corpora}, a term we borrow from natural language translation, where it refers to texts on similar topics in different languages \cite{NMTComparableCorpora,low_resource_comp_corpora}.
Here, we use it to refer to programs with similar functionality in different languages \cite{avatar,xcodeeval}.
As shown in \autoref{example_intro}, although the programs paired in a comparable example may have different algorithms and structures (e.g., global code vs.\ class functions), they are likely to have similar constructs and may even have some lines that can be matched.

\edit{In this paper, we \emph{study what the model learns from comparable corpora} by building three types of comparable examples:
(1) \emph{Naturally existing}, where we leverage independently-written solutions of the same coding problem;
(2) \emph{Generated}, where we collect programs with docstrings in one language and apply a code generation model to generate programs in another language; and
(3) \emph{Retrieved}, where we either retrieve a program's k nearest neighbors (KNN) or simply choose a random program in another language.
Among them, (1) contains cleaner examples, which are guaranteed to be bug-free. (3) covers programs from a larger variety of sources, providing more diverse training signals.}

\edit{In addition to the functional equivalence between languages, the model should also learn the equivalence between different programs in the target language.} This is challenging due to limited finetuning data. Furthermore, the majority of finetuning data only have one reference translation (e.g., 82.5\% in AVATAR, \citealt{avatar}), which is likely to cause overfitting to a single translation without fully capturing its semantics.

As a result, in this paper, we \emph{generate multiple translation references} for the finetuning data.
Specifically, after training on comparable corpora, we finetune a model on the original parallel data, generate multiple translations for each example, and use automatically generated test cases to filter out incorrect translations.
\edit{By training on different functionally equivalent programs, we reduce overfitting to a single output and improve the modeling of the target language.}

Combining the two techniques, we name our full approach \ours, a code \textbf{Trans}lation model trained with \textbf{C}omparable corpora and \textbf{M}ultiple references.
\edit{Extensive experiments show that \ours significantly improves CodeT5 \cite{codet5}, which is initialized from the same pretrained checkpoint, for an average of 7.5\% Computational Accuracy (CA@1) over translation between 6 language pairs.}
\ours also significantly outperforms the state-of-the-art method in 5 out of 6 language pairs, while reaching parity on the other one.

\edit{Analyses of our two techniques suggest that:
(1) All three types of comparable corpora (including random program pairs) improve the syntax accuracy and perplexity of the translation outputs and lead to better final performance.
(2) Both naturally existing and generated comparable corpora help the model generate constructs that match the input. The combination of them gives the largest performance gain.
(3) By training with multiple references, the model generates more unique correct translations within a certain budget, which indicates better functional equivalence of the target language is learned.}

\start{Contributions}
\edit{(1) We build and study three types of comparable corpora for code translation.
(2) We improve the modeling of target language} by generating, verifying, and selecting additional references for existing parallel translation data.
(3) We demonstrate that our model significantly outperforms state-of-the-art methods on 5 language pairs.
\edit{Our analyses provide insights into what the model learns in code translation for future researchers.}

\section{Related Work}
\start{Code Translation}
Previous work has tackled the problem of translating code written in one programming language to another. 
\citet{Karaivanov2014PhraseBasedST}, \citet{Nguyen2013LexicalSM,Nguyen2015DivideandConquerAF}, \citet{Phan2017StatisticalMO,Oda2015LearningTG} applied statistical machine translation techniques to code translation, while \citet{TreetotreeNN} introduced a tree-to-tree neural translation approach. 
Further improvements were achieved by pre-trained language models of code such as CodeBERT \cite{codebert}, PLBART \cite{plbart}, and CodeT5 \cite{codet5}. 
However, the above approaches require finetuning on parallel data, which is often scarce. 

\start{Data Scarcity in Code Translation}
To tackle the data scarcity issue, TransCoder \cite{transcoder} uses back translation for unsupervised code translation.
DOBF \cite{dobf}, TransCoder-ST \cite{transcoderST}, and S\&G \cite{SandG} respectively improve TransCoder with de-obfuscation pre-training, self-training, and pairing up the model with a generation and a summarization model.
However, the best-performing approach, TransCoder-ST, is only able to generate parallel data for standalone functions where the model can already generate a correct solution within a limited budget.
In contrast, the comparable corpora we use to train \ours include code with arbitrary structure and content.
\ours also has much better efficiency, as self-training requires running a much larger number of test cases than generating multiple references.
\edit{MultiPL-E \cite{multiple} also automatically generates test cases, but retries until the output passes all test cases. We do not directly compare to it since it requires multiple rounds of translation.}

\start{Data Augmentation for Natural Language Translation}
The lack of parallel training data is a fundamental problem in the field of translation, which has led the NLP community to develop several data augmentation techniques in response. 
\textit{Comparable Corpora}, as defined by \citet{first_comp_corpora}, ``are texts that, while not parallel in the strict sense, are somewhat related and convey overlapping information''. 
\citet{methods_collect_comp_corpora} and \citet{harvesting_comp_corpora} present methods for collecting comparable corpora to study when and how to use them. 
\citet{handle_with_care_comp_corpora} and \citet{NMTComparableCorpora} identify information balance, alignment, and length differences between source and target as key factors affecting translation quality. 
In this work, we extend the study of comparable corpora to code translation. 
We show that code translation can benefit from multiple types of comparable corpora even if there is already high-quality parallel data. 
We also provide analyses on why comparable corpora are beneficial.

Another data augmentation strategy is using multiple translation references. 
\citet{truly_exploring_multi_ref} and \citet{simulated_multi_ref} found that using multiple references can be beneficial in low-resource settings.
An advantage of working with code is that test cases can be used to filter model-generated translated references to ensure functional equivalence to the source, which we exploit in \ours.

\begin{figure*}[ht]
    \centering
    \includegraphics[width=\linewidth]{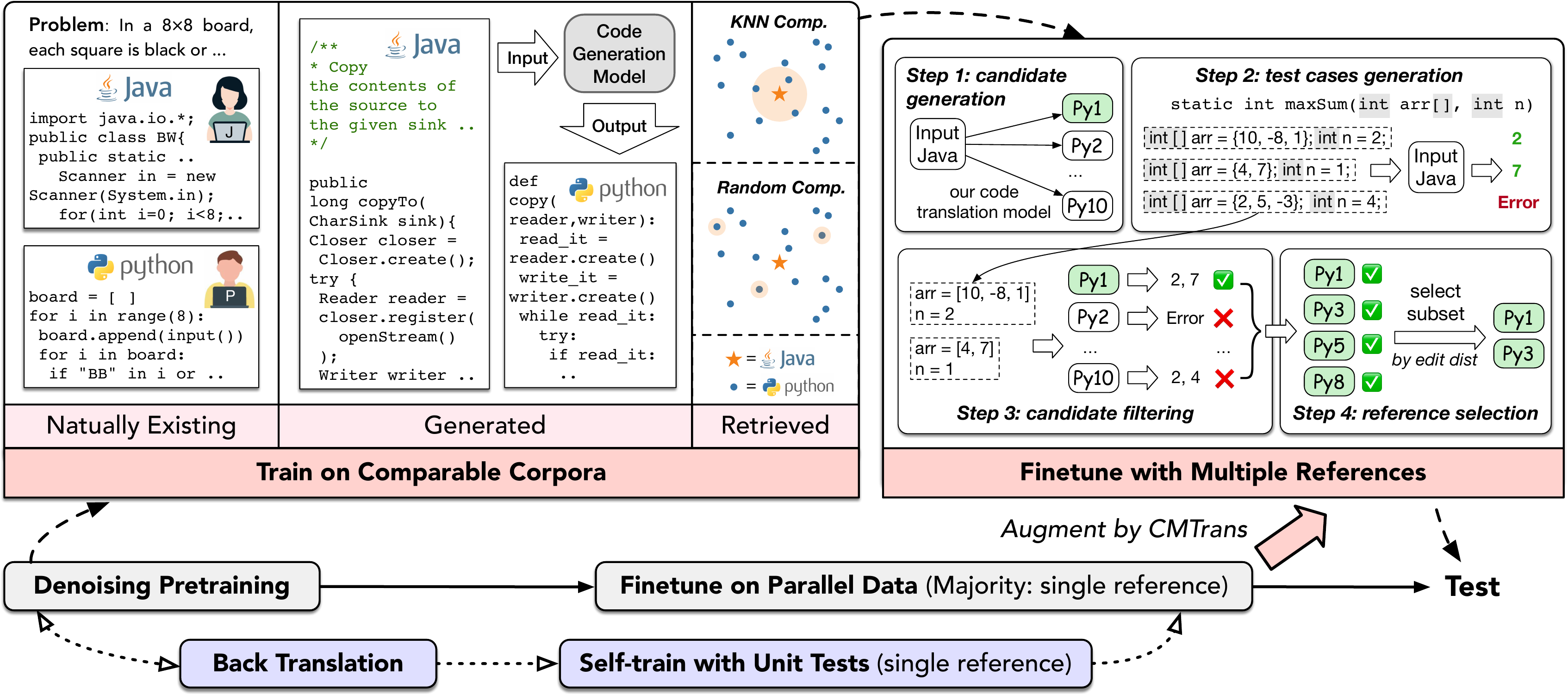}
    \upv
    \vspace{-0.5em}
    \caption{An example of \ours for Java-to-Python translation. We compare the pipeline of \hlours{diagramRed} to the \ctext{diagramGrey}{standard pipeline} of code translation (e.g., finetuned CodeT5, \citealt{codet5}) and the \ctext{diagramPurple}{self-supervision-and-fine-tuning method} of TransCoder-ST (\citealt{transcoderST}).}
    \label{method}
    \downv
\end{figure*}

\section{Methodology}
In this section, we introduce our data augmentation method that trains the model first on comparable corpora, which are either naturally existing, generated, or retrieved (Section~\ref{sec:comparable}) and then on parallel data with additional references generated by our model (Section \ref{sec:multiref}).

\subsection{Problem Formulation}
We formulate code translation as a sequence-to-sequence (Seq2Seq) generation problem following existing work \cite{codexglue,avatar,transcoder}.
The input is the code in the source language $S = \{s_1, s_2, ..., s_n \}$, which is a sequence of tokens.
We apply a model to translate the input code into the target language $T = \mathcal{M}_{\theta}(S) = \{t_1, t_2, ..., t_m \}$, where the model $\mathcal{M}_{\theta}$ is typically finetuned for code translation. 
Alternatively, the model may generate $k$ candidate outputs for each input: $\mathcal{M}_{\theta}(S,k) = \mathbb{T} = \{ T^{(1)}, T^{(2)}, ..., T^{(k)} \}$. 

\subsection{Training on Comparable Corpora}
\label{sec:comparable}

\edit{As shown in \autoref{method}, we study three types of comparable corpora: naturally existing, generated, and retrieved ones. Here we introduce how we build each type of comparable examples and train our model on them.}
We provide details of the comparable corpora datasets we use or build in Section \ref{sec:setup} and provide examples in Appendix \ref{sec:app_case_cc}.

\start{Naturally Existing Comparable Examples}
We make use of comparable corpora collected from programming contests by existing datasets \cite{avatar,xcodeeval,codenet}, which mainly consist of solutions to the same contest problems written in different languages.
These comparable examples are typically confined to specific domains such as dynamic programming or graph theory.

\start{Generated Comparable Examples}
To cover programs in more diverse domains, we present a method that automatically generates comparable examples using natural-language-to-code generation models.

Specifically, we leverage a monolingual corpus of functions with natural language documentation (i.e., docstrings) to describe their functionality.
In our experiments, we use GitHub functions with docstrings extracted by CodeSearchNet \cite{codesearchnet}.
For each function $S_c$ in the corpus, we feed the natural language documentation to a code generation model finetuned in the other language, which is trained to generate a program based on the given natural language description.
Similarly to code translation, the code generation model can generate multiple candidate outputs $\{ T_c^{(1)}, T_c^{(2)}, ..., T_c^{(k)} \}$.
We then select the one with the highest probability, which is paired up with the extracted program as a comparable example.

\start{Retrieved Comparable Examples}
\edit{To study the effect of the quality of comparable corpora, we further build \textbf{KNN Comparable Corpora}.
We compute the embedding of all the programs in the source and target language in the dataset using a finetuned code retrieval model~\cite{codesearchnet}.
For each program $S$ in the source language with embedding $emb(S)$, we retrieve its k nearest neighbors in the target language by the cosine similarity between their embeddings: $sim(S,T) = \langle emb(S), emb(T) \rangle$.}

\edit{Finally, we build \textbf{Random Comparable Corpora} by pairing random programs in the source and target language.
In principle, such program pairs do not contain any information on functional equivalence and allow us to better understand whether comparable corpora can improve the modeling of the target language and hence improve the translation quality.}

\begin{table*}[htb]
\centering
\resizebox{\textwidth}{!}{
\begin{tabular}{lcccc|c}
\toprule
\multirow{2}{*}{\bf Dataset $\rightarrow$}
& \multicolumn{3}{c}{\textbf{Train (Comparable)}} & \textbf{Train (Parallel)} & \textbf{Test} \\ 
\cmidrule(lr){2-4}  \cmidrule(lr){5-5} \cmidrule(lr){6-6}
& \textbf{Gen-comp} & \textbf{XCodeEval} & \textbf{AVATAR-comp} 
& \textbf{AVATAR-para} & \textbf{TransCoder-test} \\
\midrule
\bf \# C++ $\leftrightarrow$ Java & 4,053* & 3,414 & -- & 3,226* & 482 C-to-J, 467 J-to-C \\
\bf \# C++ $\leftrightarrow$ Python & 4,053* & 4,376 & -- & 3,226* & 464 C-to-P, 467 P-to-C \\
\bf \# Java $\leftrightarrow$ Python & 22,181* & -- & 5,937 & 3,391 & 464 J-to-P, 482 P-to-J \\ 
\bf Source 
& \begin{tabular}[c]{@{}c@{}}
Github (Java $\leftrightarrow$ Python) \\
AIZU, AtCoder (Others)
\end{tabular}
& Codeforces 
& \begin{tabular}[c]{@{}c@{}} 
AtCoder, Codeforces, ProjectEuler, \\ 
CodeJam, GeeksforGeeks, LeetCode, 
\end{tabular} 
& GeeksforGeeks 
& GeeksforGeeks \\
\bottomrule
\end{tabular}
}
\caption{Number of problems per dataset. ``Gen-comp'' is the comparable corpora dataset we generate. C-to-J, C-to-P, ... denote the test examples of C++-to-Java, C++-to-Python translation, etc. * denotes data we build.}
\label{tab:dataset}
\end{table*}

\start{Model Training}
After constructing a corpus of comparable examples $\mathcal{D}_c = \{(S_c, T_c)\}$, we input the program in the source language $S_c$ into our model $\mathcal{M}_{\theta}$ and maximize the log-likelihood of its corresponding program in the target language $T_c = \{t_{c,1}, t_{c,2}, ..., t_{c,m} \}$:
\begin{align}
\begin{split}
&\mathcal{L}_{M T}(\mathcal{D}_c) = \sum_{(S_c, T_c) \in \mathcal{D}_c} P_{\theta}(T_c | S_c) \\
&P_{\theta}(T_c | S_c) = -\sum_i \log \left(P_{\theta}\left(t_{c,i} | S_c, t_{c,1} \ldots t_{c,i-1} \right)\right)
\end{split}
\label{eq:mtloss}
\end{align}

Here $\theta$ is the parameters of $\mathcal{M}_{\theta}$, which is initialized from an encoder-decoder model pretrained on code \cite{codet5}.
After training the model on the corpus comparable examples $\mathcal{D}_c$ till convergence, we finetune it on the dataset of parallel examples $\mathcal{D}_p$, where the loss $\mathcal{L}_{M T}(\mathcal{D}_p)$ is computed by \autoref{eq:mtloss} as well.

By maximizing the probability of the target $T_c$, the model will learn to generate fluent code in the target language.
Furthermore, in general, programs in a comparable example often exhibit the same types of constructs.
In the examples in \autoref{example_intro}, to check whether a board is valid, no matter what algorithm is used, the program will always need a loop to read the board and use if statements for the validity check.
As a result, in principle, the model can also learn to generate the same types of constructs as the input program $S_c$, which is beneficial for generating accurate translations.

\subsection{Finetuning with Multiple References}
\label{sec:multiref}
In addition to comparable corpora, we also provide our model with more diverse training signals by finetuning with multiple references. 
By providing the model with programs with the same functionality, we encourage the model to learn a better representation space for the target language and hence benefit the translation.

Since the majority of source programs only have one reference in existing datasets of parallel examples \cite{avatar}, we apply a series of steps to generate additional references, which are illustrated in \autoref{method}.
The first step is to finetune a model with the original parallel data, and then use the finetuned model to generate multiple translation candidates $\{ T_c^{(1)}, T_c^{(2)}, ..., T_c^{(k)} \}$ for each source program in the parallel data.

In the second and third steps, similar to TransCoder-ST \cite{transcoderST}, we leverage automatically generated unit tests to filter out candidates with different behaviors as the source program.
Specifically, we extract the input arguments of the source programs, randomly produce a set of test inputs based on the argument types, and feed the test inputs to the source program.
We filter out test inputs that cause compilation or runtime errors.
Finally, we feed the remaining test inputs to each candidate and only keep the ones that have exactly the same output as the source program.

Notice that some of the translations may only have small differences (e.g., \texttt{i++} vs. \texttt{i+=1}). To obtain a diverse subset of references, we select the most distinct $k$ translations for each source program by their string edit distance.
These $k$ translations are added as additional references to the finetuning set.

\begin{table*}[htb]
\centering
\resizebox{0.9\textwidth}{!}{
\begin{tabular}{lccccccccc}
\toprule
\multirow{2}{*}{\bf Model $\downarrow$}
& \multicolumn{3}{c}{\bf Java-to-Python} 
& \multicolumn{3}{c}{\bf Python-to-Java} 
& \multicolumn{3}{c}{\bf Avg of 6 Pairs} 
\\
\cmidrule(lr){2-4}  \cmidrule(lr){5-7} \cmidrule(lr){8-10}
& \bf BLEU & \bf CB & \bf CA@1 
& \bf BLEU & \bf CB & \bf CA@1 
& \bf BLEU & \bf CB & \bf CA@1 \\
\midrule
TransCoder \cite{transcoder} & 72.4 & 67.9 & 49.1 & 65.4 & 70.7 & 35.7 
& 72.0 & 75.0 & 51.7 \\
DOBF \cite{dobf} & 72.2 & 67.5 & 52.2 & 67.7 & 71.2 & 44.4 
& --- & --- & --- \\
TransCoder-ST \cite{transcoderST} & 73.1 & 68.7 & 68.5 & 70.0 & 71.9 & 58.1 
& 71.3 & 74.9 & 66.3 \\
\midrule
CodeBERT \cite{codebert} & 52.0 & 48.9 & 10.4 & 45.4 & 45.0 & 4.2 
& --- & --- & --- \\
CodeT5 \cite{codet5} & 79.4 & 72.5 & 61.0 & 79.0 & 75.9 & 52.7 
& \underline{83.6} & 80.0 & 62.6 \\
PLBART \cite{plbart} & \underline{79.9} & \underline{73.2} & 68.9 & 80.5 & 76.8 & 57.5 
& --- & --- & --- \\
TransCoder-ST-ft \cite{transcoderST} & 79.3 & 72.9 & \underline{69.4} & \underline{81.4} & \underline{78.4} & \underline{62.0} 
& 81.8 & \underline{80.2} & \underline{67.6} \\
\midrule
\ours & \bf 80.1 & \bf 74.2 & \bf 73.5 & \bf 84.3 & \bf 82.1 & \bf 66.0 
& \bf 84.9 & \bf 82.0 & \bf 70.1 \\ 
\bottomrule
\end{tabular}
}
\caption{Java-Python translation results on TransCoder-test. We copy the results of all the baselines reported by \citet{avatar}. 
CB and CA@1 stand for CodeBLEU and Computational Accuracy. 
We highlight the \textbf{best} results under each metric with Bold and underline the \underline{second-best} results.}
\label{tab:main}
\end{table*}

\begin{table*}[htb]
\centering
\resizebox{\textwidth}{!}{
\begin{tabular}{lcccccccccccc}
\toprule
\multirow{2}{*}{\bf Model $\downarrow$}
& \multicolumn{3}{c}{\bf C++-to-Java} 
& \multicolumn{3}{c}{\bf C++-to-Python}
& \multicolumn{3}{c}{\bf Java-to-C++} 
& \multicolumn{3}{c}{\bf Python-to-C++} \\
\cmidrule(lr){2-4}  \cmidrule(lr){5-7} \cmidrule(lr){8-10} \cmidrule(lr){11-13}
& \bf BLEU & \bf CB & \bf CA@1 
& \bf BLEU & \bf CB & \bf CA@1 
& \bf BLEU & \bf CB & \bf CA@1 
& \bf BLEU & \bf CB & \bf CA@1 \\
\midrule
TransCoder \cite{transcoder} & 84.0 & 86.7 & 65.1 & 75.2 & 73.4 & 47.1 & 83.6 & 85.4 & 79.8 & 51.6 & 65.7 & 32.6 \\
TransCoder-ST \cite{transcoderST} & 78.8 & 85.2 & 68.0 & 73.1 & 73.0 & 61.3 & 76.7 & 83.7 & \bf 84.6 & 55.8 & 67.1 & 56.7 \\
\midrule
CodeT5 \cite{codet5} & \underline{90.9} & 90.0 & 65.1 & \underline{82.9} & \underline{75.4} & 56.5 & 89.1 & \bf 88.5 & 81.6 & \underline{79.8} & \underline{77.9} & 58.5 \\
TransCoder-ST-ft \cite{transcoderST} & 88.7 & \underline{90.1} & \underline{68.3} & 75.4 & 74.3 & \underline{62.5} & \underline{89.3} & 87.8 & \bf 84.6 & 76.7 & 77.6 & \underline{59.1} \\
\midrule
\ours & \bf 91.6 & \bf 90.5 & \bf 71.4 & \bf 83.7 & \bf 77.2 & \bf 64.2 & \bf 89.7 & \underline{88.4} & \underline{84.4} & \bf 82.0 & \bf 79.3 & \bf 61.2 \\
\bottomrule
\end{tabular}
}
\caption{Java-C++ and Python-C++ translation results on TransCoder-test. The CA@1 results of TransCoder and TransCoder-ST are copied from the TransCoder-ST paper. We evaluate BLEU and CodeBLEU using their released checkpoints. We finetune and evaluate CodeT5 and TransCoder-ST-ft on our own.}
\label{tab:cpp}
\end{table*}

\section{Experiments}
In this section, we conduct experiments to answer four
research questions: 
(\textbf{RQ1}) How do \ours and its ablations perform compared against state-of-the-art approaches for code translation?
(\textbf{RQ2}) What can the model learn from comparable corpora?
(\textbf{RQ3}) What can the model learn from multiple references? 
(\textbf{RQ4}) How is \ours affected by the size of comparable corpora and the number of references?

\subsection{Experimental Setup}
\label{sec:setup}
We initialize \ours from the pretrained checkpoint of CodeT5 \cite{codet5}, an encoder-decoder model pretrained on code files from varied languages with denoising objectives.

\start{Datasets}
We list the dataset statistics in \autoref{tab:dataset}.
All the methods are evaluated on the TransCoder-test dataset~\cite{transcoder}.
We train our method on xCodeEval \cite{xcodeeval}, a comparable corpora dataset, and AVATAR \cite{avatar}, which contain both comparable corpora and parallel functions (denoted as AVATAR-comp and AVATAR-para, respectively).
All the supervised baseline methods are finetuned on AVATAR-para before evaluation.

Since AVATAR-para does not contain C++ functions, we add a parallel C++ function for each training example in AVATAR-para. Specifically, we generate 50 C++ translations for each Java function by TransCoder-ST, TransCoder, and finetuned CodeT5 and filter the translations with unit tests.

\start{Construction of Comparable Corpora}
\edit{
We conduct the study on different types of comparable corpora (\CC) on Java $\leftrightarrow$ Python translation.
We use AVATAR-comp\footnote{There are two versions of AVATAR-comp with slight differences. We use the first version because most of our experiments were finished before the second version was released.} as the naturally existing \CC.
KNN and random \CC are also retrieved from AVATAR-comp.
To build the dataset of generated \CC (denoted as Gen-Comp), we generate from functions with docstrings in the CodeSearchNet~\cite{codesearchnet} dataset.}

We train \ours first on the best combination of comparable corpora, which is natural and generated \CC, and then finetuned on AVATAR-para.
For language pairs other than Java $\leftrightarrow$ Python, we use xCodeEval as naturally existing \CC and generate \CC from CodeNet~\cite{codenet}. More details can be found in Appendix~\ref{sec:implementation}.

\start{Evaluation metrics}
Our primary metric is the Computational Accuracy (CA@k), which evaluates whether at least 1 out of k translation candidates generates the same outputs as the reference for all the given test inputs.
Following previous work \cite{avatar}, we report \textbf{CA@1} results and also report \textbf{BLEU}, which computes the overlap between candidate and reference translations \cite{BLEU}, and \textbf{CodeBLEU}: the weighted average of token level match, syntax level match, and Dataflow match \cite{codebleu}.

Appendix \ref{sec:implementation} contains more details about the implementation and baselines.

\subsection{Main results}
\autoref{tab:main} and \autoref{tab:cpp} show the code translation results on TransCoder-test.

\ours substantially outperforms CodeT5, which is initialized from the same pretrained checkpoint and finetuned with the original parallel data, by an average improvement of 7.5\% CA@1.
\ours also significantly outperforms the state-of-the-art methods, TransCoder-ST-ft, on 5 out of 6 language pairs, and reaches parity on Java-to-C++ translation.
Note that TransCoder-ST-ft generates test cases for 103,488 functions and executes these test cases over 4 iterations of training. 
In comparison, our method only generates test cases for 3,391 functions and executes the test cases once, resulting in better efficiency.

Our results show that the advantage of \ours is larger on translations between Java and Python. The reason may be that the parallel data we generate for translations involving C++ are only verified by automatically generated test cases, which might not cover all the boundary cases and could introduce noise to the finetuning set.

\edit{Following previous work~\cite{transcoderST,avatar}, we report the results of one checkpoint. 
We also conduct the t-test in Appendix~\ref{sec:ttest}, which indicates that \ours significantly outperforms the best baseline over CA@1 with p-value $<$ 0.01 for 4 out of 6 language pairs and with p-value $<$ 0.05 for one language pair.}

\subsection{Performance Analysis}
\start{Ablation studies}
\autoref{tab:ablation} presents the ablations of our approach. 
\edit{
We first study the effectiveness of each type of comparable corpora and some combinations of them. All the ``+ \CC'' ablations denote training CodeT5 on comparable corpora and then finetuneing on AVATAR-para.
We also ablate ``+ Multi-Ref'', which is finetuned on multiple references directly after pretraining.
The results show that both Multi-Ref and the best \CC provide a large performance gain and stacking them (the full \ours) has the best performance.}

\edit{
As for the effectiveness of different \CC, we can see that code translation can benefit from all these types of \CC. The combination of Natural and generated \CC has the best performance. The reason might be both \CC have relatively high data quality and further adding the other two \CC in training introduces noise to the model and hinders the learning of functional equivalence.}

Notice that with the same finetuning set, \textit{\CC\ Only} still outperforms TransCoder-ST-ft, which indicates that compared to self-training, training on comparable corpora has not only better efficiency but better effectiveness.
We hypothesize that the reason may be that the comparable corpora contain more diverse structures and contents. This provides more diverse training signals to the model and hence improves the generalization.

\begin{table}[t]
\centering
\resizebox{\linewidth}{!}{
\begin{tabular}{lcc}
\toprule
\bf Model $\downarrow$ & \bf J-to-P CA@1 & \bf P-to-J CA@1 \\
\midrule
CodeT5 & 61.0 & 52.0 \\
+ Random \CC & 64.7 & 54.6 \\
+ KNN \CC & 63.8 & 55.4 \\
+ Generated \CC & 64.0 & 63.1 \\
+ Natural \CC & 68.8 & 62.9 \\ 
\midrule
+ All \CC & 62.5 & 55.4 \\
+ Natural \& Generated & 70.0 & 64.1 \\ 
+ Multi-Ref & 68.1 & 61.6 \\ 
\ours & \bf 73.5 & \bf 66.0 \\ 
\bottomrule
\end{tabular}
}
\caption{Ablation studies of \ours. We report CA@1 results. J-to-P and P-to-J stand for Java-to-Python and Python-to-Java results. We put the experiments on each type of comparable corpora in an individual block.}
\label{tab:ablation}
\end{table}

\start{Translation with Limited Parallel Data}
To analyze how well our method tackles the data scarcity challenge, we compare the performance of CodeT5 and \ours when finetuned on parallel data with different sizes.
\edit{For simplicity, we use \ours (\CC Only) to denote the ``CodeT5 + Natural \& Generated'' ablation 
and use \ours (Multi-Ref\ Only) to denote ``CodeT5 + Multi-Ref''.}

\begin{figure}[t]
    \centering
    \includegraphics[width=\linewidth]{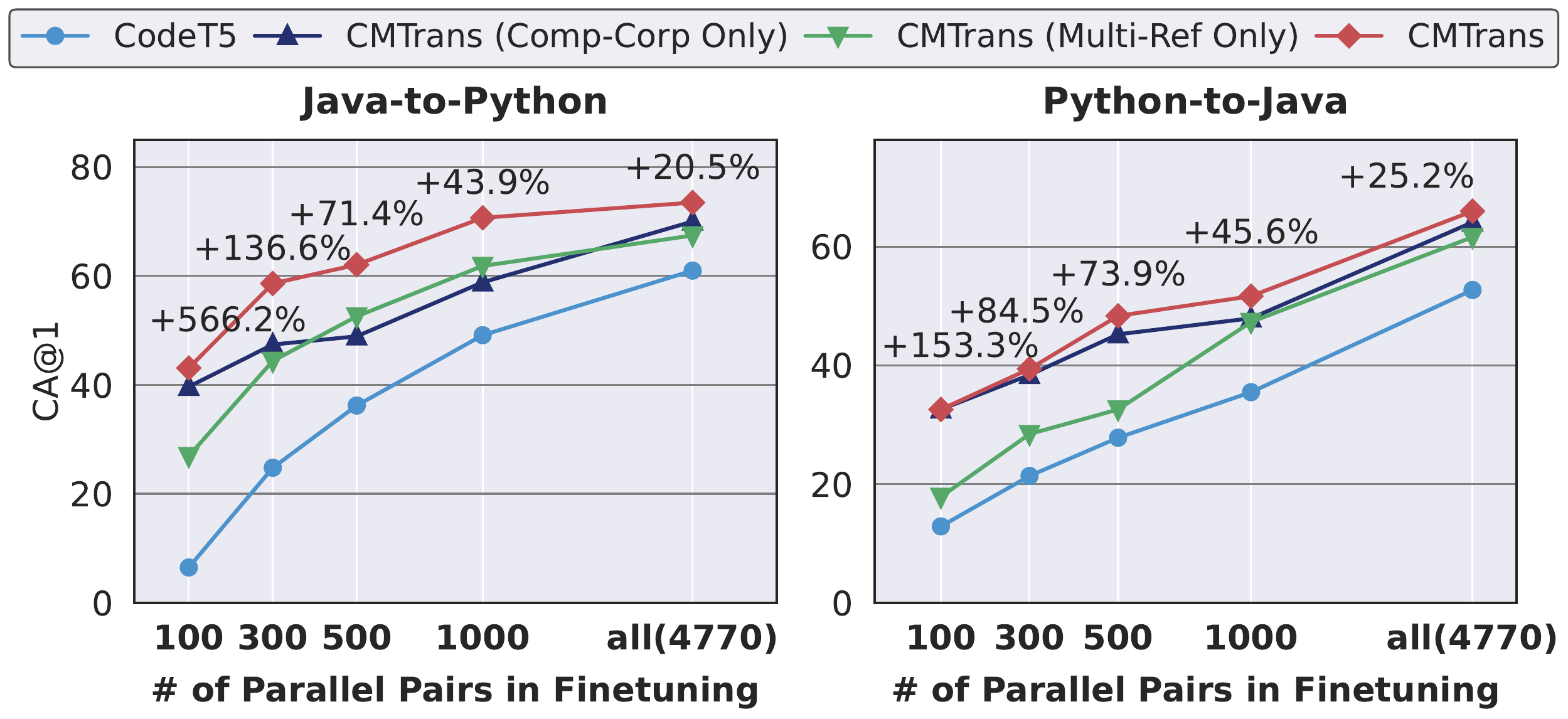}
    \caption{Translation results with different amount of parallel data. We mark the relative gain of \ours over CodeT5.}
    \label{fig:size}
\end{figure}

As shown in \autoref{fig:size}, the relative gains of \ours as well as our ablations are more pronounced when the parallel data is more limited. 
For example, when there are only 100 parallel examples for Java-to-Python translation, CodeT5 obtains 6.5 CA@1 while \ours, (\CC Only), and (Multi-Ref\ Only) obtain 43.1, 39.7, and 26.9 CA@1, respectively. 
This demonstrates the effectiveness of our two data augmentation techniques to tackle data scarcity.

\begin{figure}[t]
    \centering %
\begin{subfigure}{0.5\linewidth}
  \includegraphics[width=\linewidth]{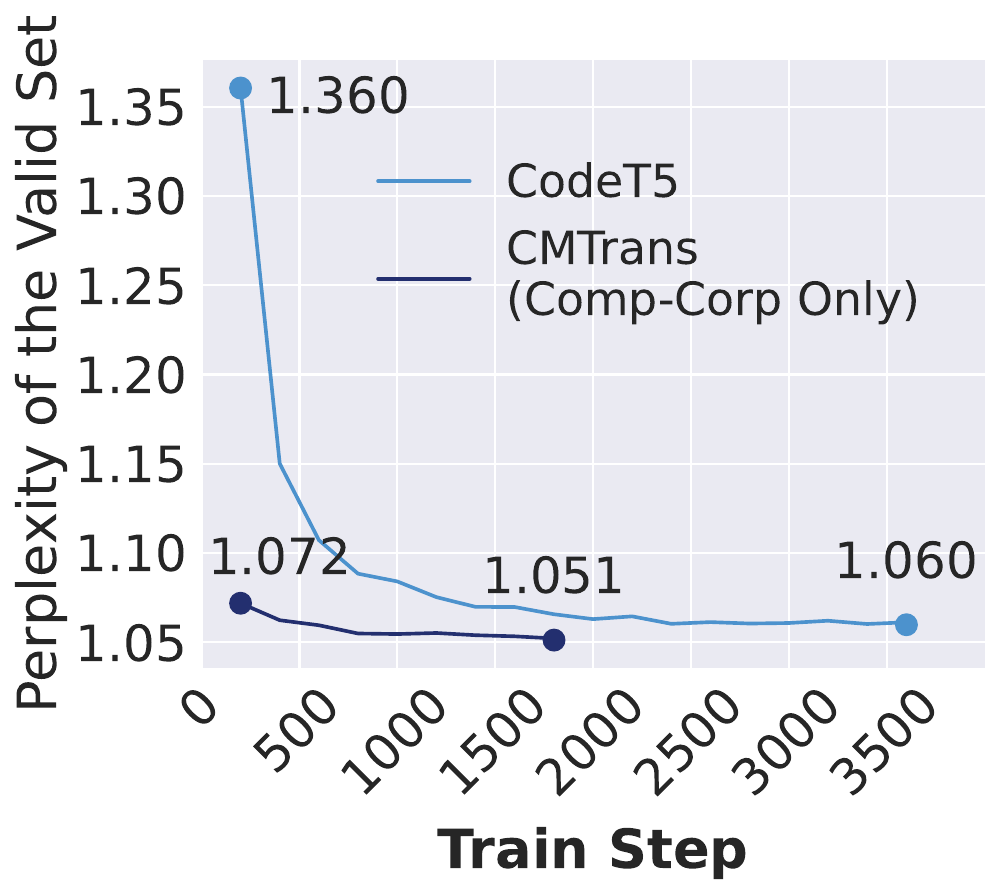}
  \caption{Java-to-Python}
  \label{fig:ppl1}
\end{subfigure}\hfil %
\begin{subfigure}{0.5\linewidth}
  \includegraphics[width=\linewidth]{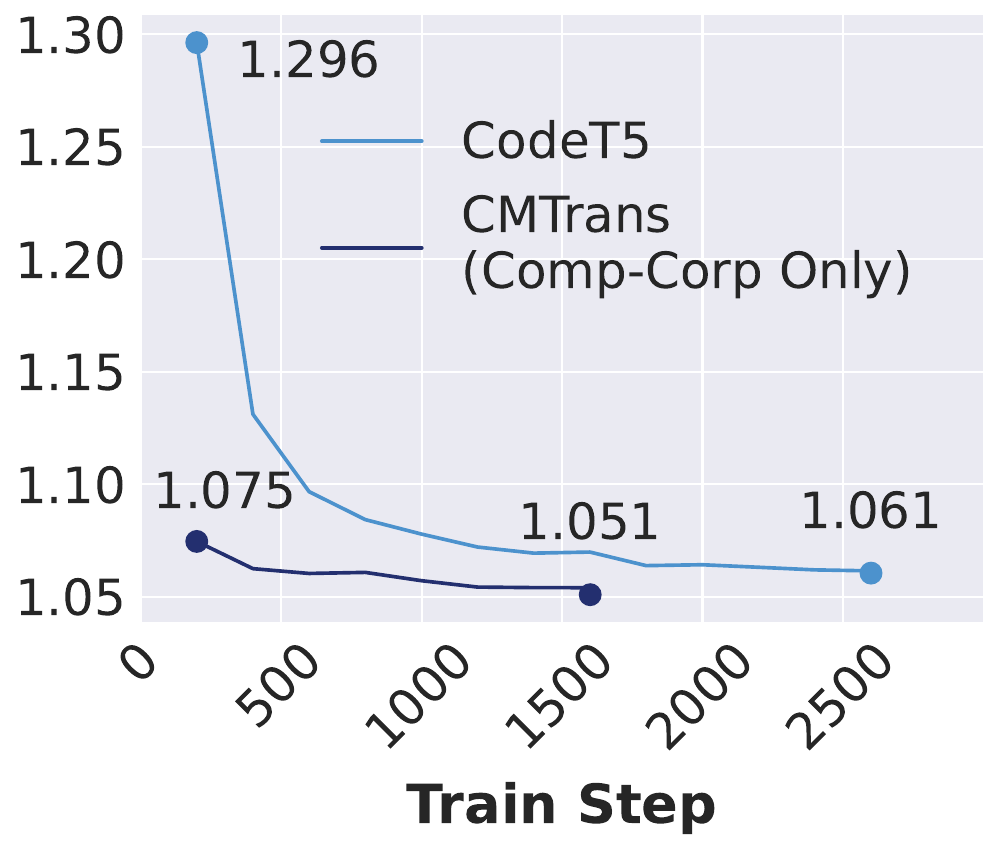}
  \caption{Python-to-Java}
  \label{fig:ppl2}
\end{subfigure}
\caption{Perplexity of validation set during finetuning.}
\label{fig:ppl}
\end{figure}
\begin{table*}[]
\centering
\resizebox{\linewidth}{!}{
\begin{tabular}{lccccccccc|ccccccccc}
\toprule
\multirow{3}{*}{\bf Model $\downarrow$}
& \multicolumn{9}{c}{\bf Java-to-Python} 
& \multicolumn{9}{c}{\bf Python-to-Java} \\
\cmidrule(lr){2-10} \cmidrule(lr){11-19} 
& \multicolumn{3}{c}{\bf LOOP} 
& \multicolumn{3}{c}{\bf IF} & \multicolumn{3}{c}{\bf ELSE IF} 
& \multicolumn{3}{c}{\bf LOOP} 
& \multicolumn{3}{c}{\bf IF} & \multicolumn{3}{c}{\bf ELSE IF} \\
\cmidrule(lr){2-4}  \cmidrule(lr){5-7} \cmidrule(lr){8-10}
\cmidrule(lr){11-13}  \cmidrule(lr){14-16} \cmidrule(lr){17-19}
& P & R & F1 & P & R & F1 & P & R & F1 
& P & R & F1 & P & R & F1 & P & R & F1 \\
 \midrule
CodeT5 (No finetune) & 100.0 & 28.5 & 44.4 & 99.1 & 35.4 & 52.2 & 0.0 & 0.0 & 0.0 &
100.0 & 14.6 & 25.5 & 100.0 & 36.6 & 53.6 & 0.0 & 0.0 & 0.0 \\
+ KNN \CC & 89.4 & 94.6 & 91.9 & 82.1 & 64.3 & 72.1 & 41.6 & 57.6 & 48.3 & 
87.9 & 98.9 & 93.1 & 90.2 & 50.9 & 65.1 & 30.5 & 16.4 & 21.3 \\
+ Generated \CC & 99.9 & 98.9 & \bf 99.4 & 99.0 & 93.2 & \bf 96.0 & 84.6 & 79.2 & \bf 81.8 & 98.3 & 96.7 & \bf 97.5 & 98.7 & 94.8 & \bf 96.7 & 83.3 & 86.4 & \bf 84.8 \\
+ Natural \CC & 95.8 & 85.4 & 90.3 & 91.2 & 73.9 & 81.6 & 50.0 & 49.6 & 49.8 & 87.4 & 98.9 & 92.8 & 88.2 & 80.4 & 84.1 & 57.8 & 33.6 & 42.5 \\
\bottomrule
\end{tabular}
}
\caption{The overlap between the types of constructs in the translation outputs and the ground truth translations. }
\label{tab:construct}
\end{table*}
\begin{table}[t]
\centering
\resizebox{\linewidth}{!}{
\begin{tabular}{lcc}
\toprule
\bf Model $\downarrow$ & \bf J-to-P SA& \bf P-to-J SA \\
\toprule
No finetuning \\
\midrule
CodeT5 \cite{codet5} & 1.1 & 1.7 \\
+ Random \CC & 20.0 & 41.3 \\
+ KNN \CC & 22.0 & \bf 56.4 \\
+ Generated \CC & \bf 41.4 & 43.2 \\
+ Natural \CC & 34.1 & 54.6 \\
\toprule
With finetuning \\
\midrule
CodeT5 \cite{codet5} & 95.3 & 69.7 \\
+ Random \CC & 96.3 & 69.7 \\
+ KNN \CC & 96.3 & 71.0 \\
+ Generated \CC & \bf 97.6 & 71.2 \\
+ Natural \CC & 97.4 & \bf 76.4 \\
\bottomrule
\end{tabular}
}
\caption{Syntax Accuracy (SA) on TransCoder-test before finetuning, which evaluates whether the program can be compiled without syntax errors.}
\label{tab:EA}
\end{table}

\subsection{Influence of Comparable Corpora}
To answer \textbf{RQ2}, we hypothesize that training on comparable corpora allows the model to produce fluent code in the target language while using similar constructs as the input.

\start{Fluency of outputs}
To validate our hypothesis, we compare the perplexity of the reference translations (in the target language) during finetuning, which reflects the fluency of outputs.
As shown in \autoref{fig:ppl}, after training on comparable corpora, the perplexity is substantially reduced before finetuning. It also converges to a lower value after finetuning on parallel data.

\edit{Similarly, \autoref{tab:EA} shows that in most scenarios, training on comparable corpora leads to fewer syntax errors both before and after finetuning, which means the programs our method generates not only have a high token-level overlap with the reference translations but are also syntactically correct.
The reason might be comparable corpora (including the random \CC) train the model to maximize the probability of a complete program in the target language, improving its language modeling ability.}

\start{Generation of matching constructs}
We observe that programs in a comparable pair typically contain similar constructs (e.g., in AVATAR-comp, for 83.41\% of Java programs with if statements, the corresponding Python programs also contain if statements).
\edit{To assess whether our model also learns to generate the correct types of constructs, for each type of construct, we consider whether the reference translation has this type of construct as the ground truth and whether the translation output has it as the predictions. Then we compute the accuracy, recall, and F1 scores.}

\edit{As shown in \autoref{tab:construct}, after training on KNN, generated, and natural \CC, the F1 score is highly improved.
The generated \CC has the highest F1 scores. The reason might be we input both the documentation and the program to the generation model, so the generated program follows the same algorithm as the input program, which results in a larger percent of comparable examples that share the same types of constructs.}

\begin{figure}[t]
    \centering %
\begin{subfigure}{0.5\linewidth}
  \includegraphics[width=\linewidth]{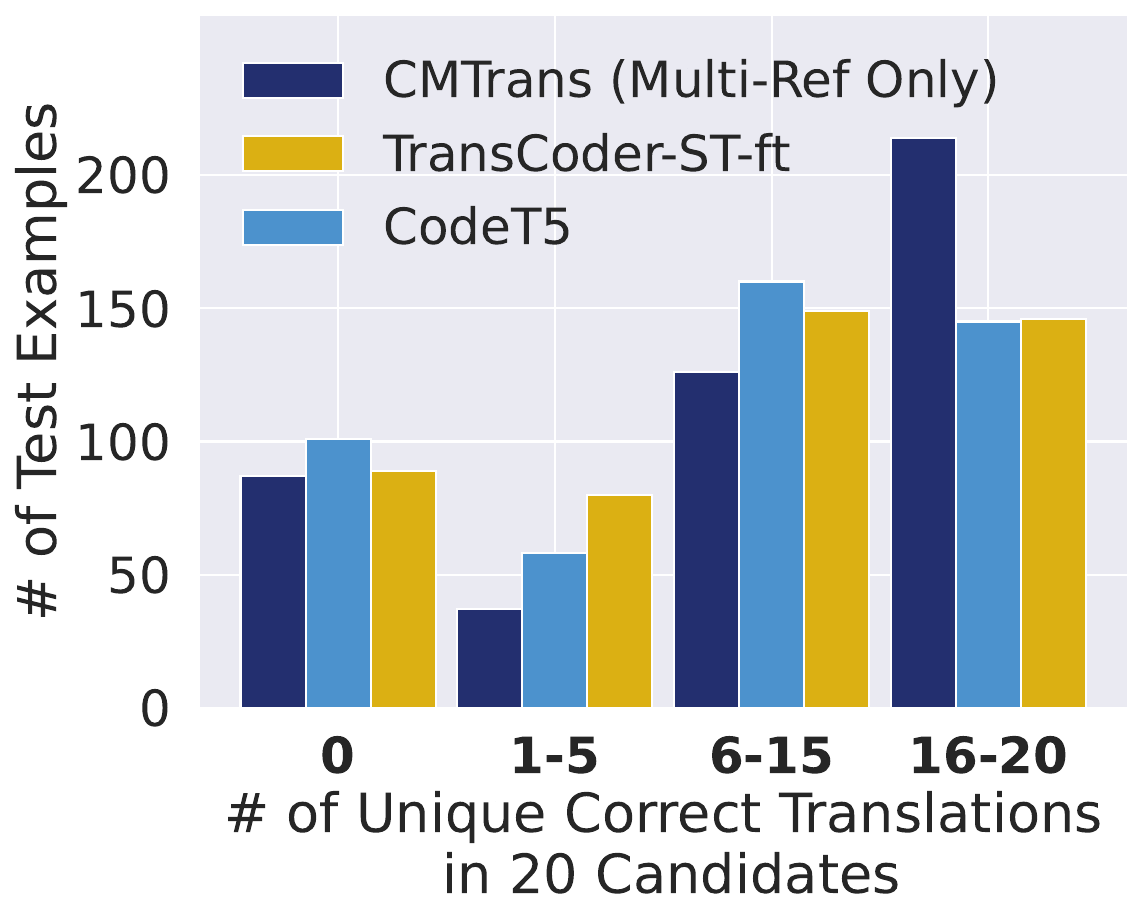}
  \caption{Java-to-Python}
  \label{fig:uni1}
\end{subfigure}\hfil %
\begin{subfigure}{0.5\linewidth}
  \includegraphics[width=\linewidth]{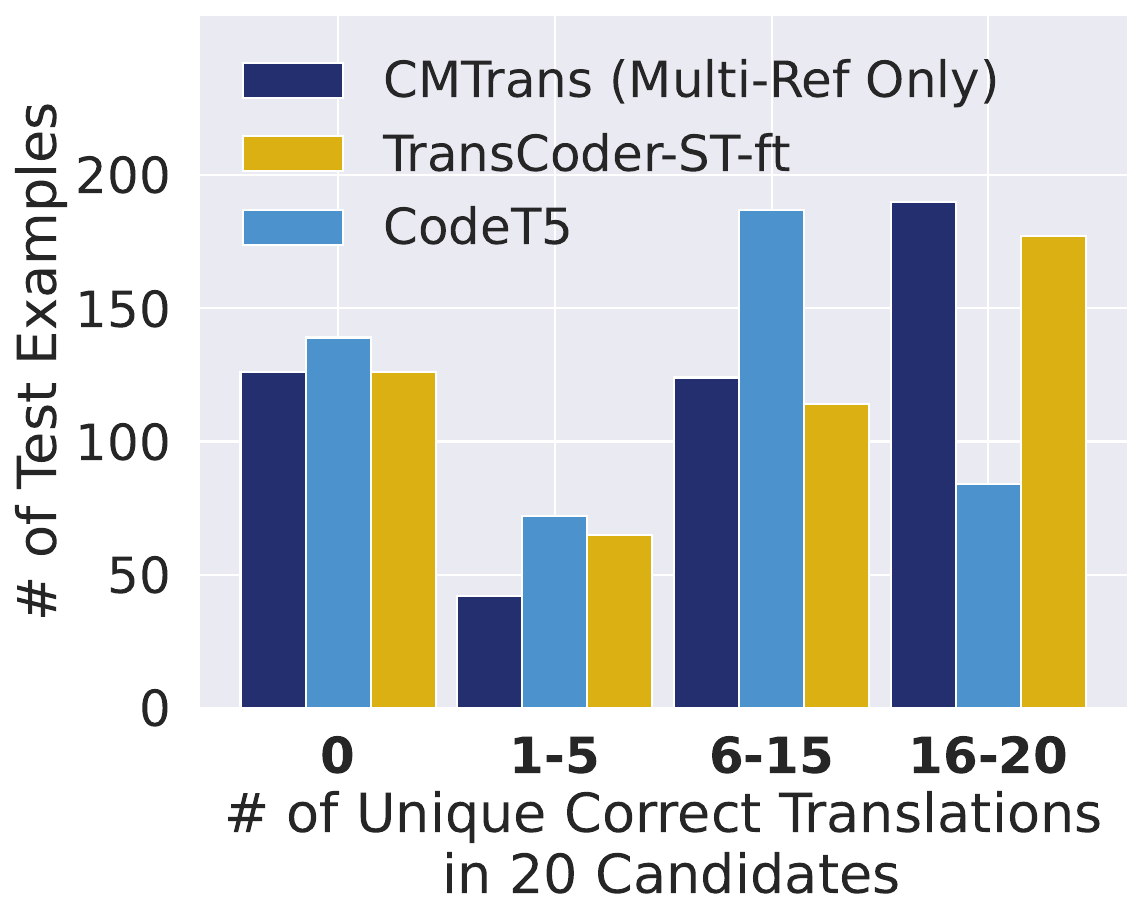}
  \caption{Python-to-Java}
  \label{fig:uni2}
\end{subfigure}
\caption{Number of unique correct translations in 20 candidates for each test example. We use beam search for each method, so the generated candidates are guaranteed to be distinct.}
\label{fig:unique}
\end{figure}

\begin{figure*}[t]
    \centering
    \includegraphics[width=\linewidth]{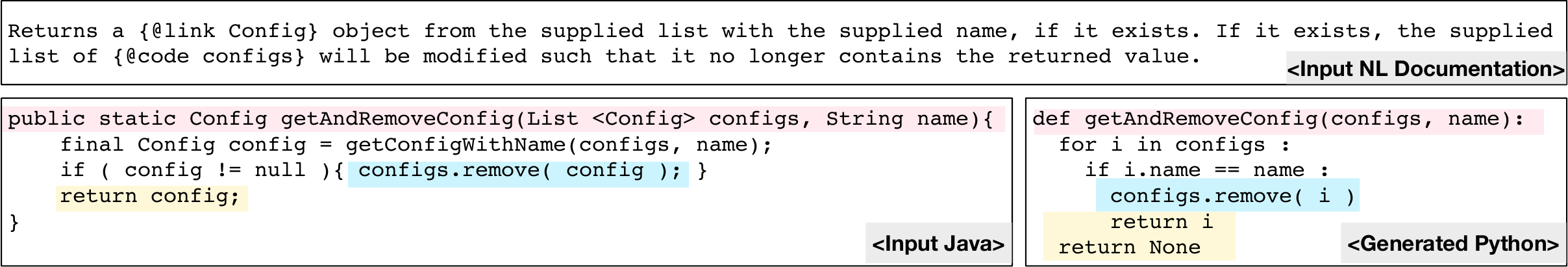}
    \caption{Case study: a comparable example generated by our method with an NL-to-code generation model. We highlight the lines in the Java program and the generated Python program that can be matched.}
    \label{fig:case}
\end{figure*}
\begin{figure*}[t]
    \centering %
\begin{subfigure}{0.5\linewidth}
  \includegraphics[width=\linewidth]{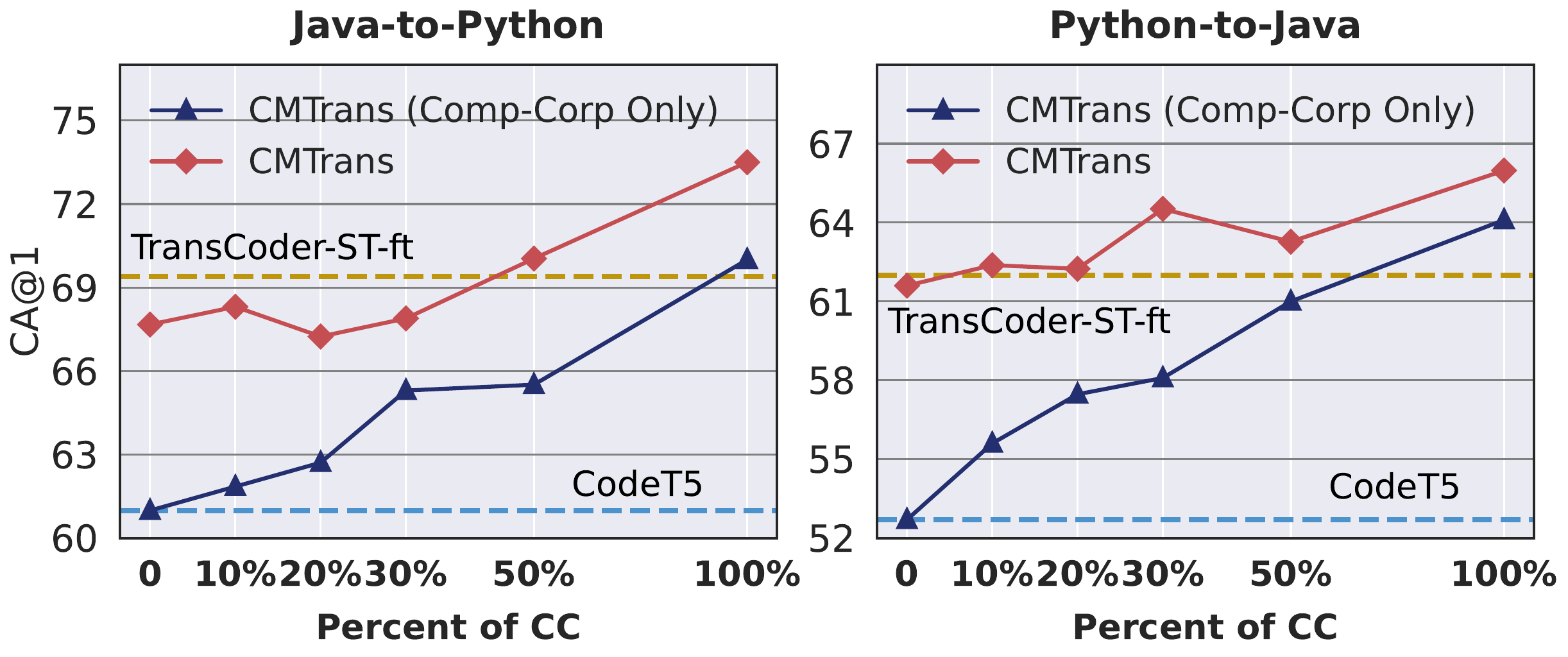}
  \caption{Effect of the size of comparable corpora.}
  \label{fig:hyper_cc}
\end{subfigure}\hfil %
\begin{subfigure}{0.5\linewidth}
  \includegraphics[width=\linewidth]{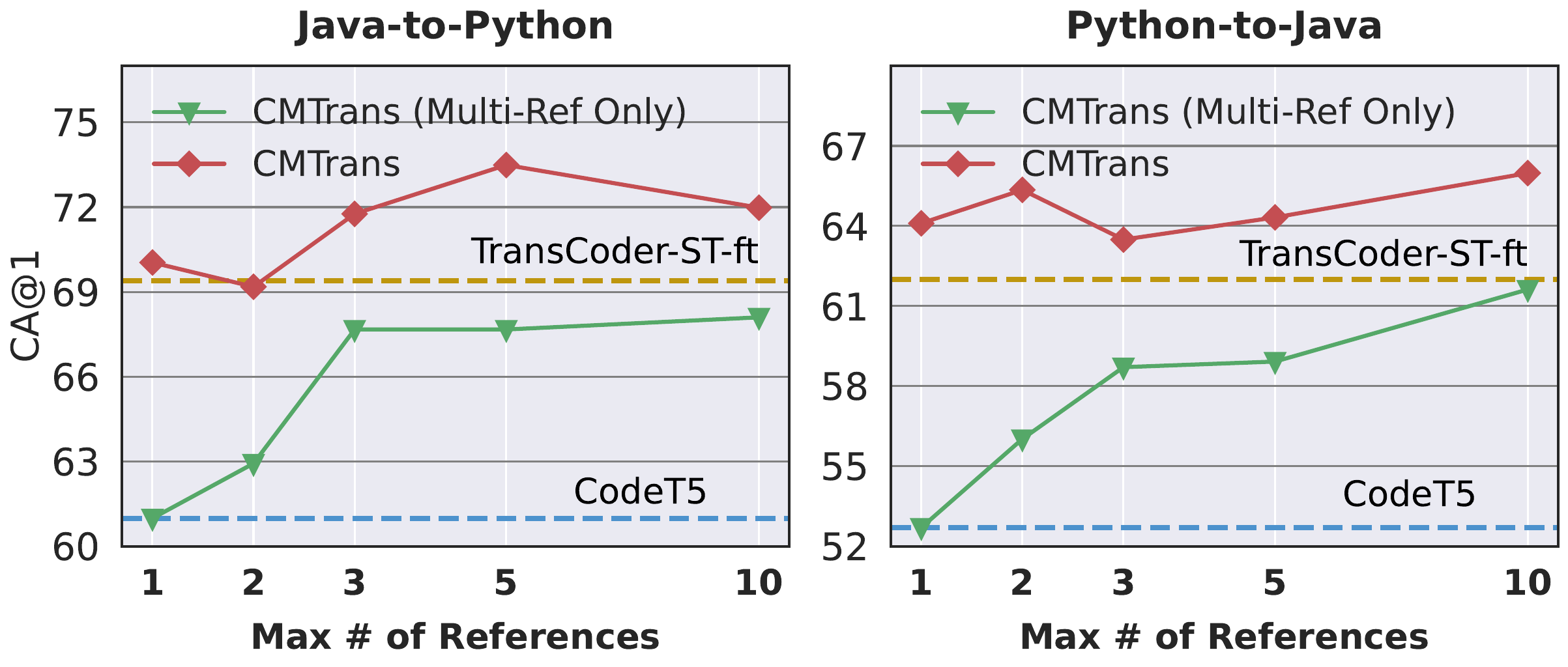}
  \caption{Effect of the max number of references for each example.}
  \label{fig:hyper_multi}
\end{subfigure}
\caption{The CA@1 score for \ours and its ablations with different hyper-parameters.}
\label{fig:hyper}
\end{figure*}

\subsection{Influence of Multiple References}
As for \textbf{RQ3}, we hypothesize that training on additional references can reduce overfitting to a single translation. 
To validate this hypothesis, we show in \autoref{fig:unique} that when trained with multiple references, our model can generate more unique correct translations using beam search within the same number of candidate outputs.
For instance, in Java-to-Python translation, there are 214 test examples where our model generates $\geq$ 16 unique translations, while there are only 145 and 146 examples for CodeT5 and TransCoder-ST-ft, respectively.
Furthermore, the scores under the ``0'' group indicate that after training with multiple references, the model generates at least one correct translation for more test examples.
In other words, in addition to CA@1, training CodeT5 on multiple references also improves on CA@20.

\subsection{Hyper-parameter analysis}
To answer \textbf{RQ4}, we analyze two hyper-parameters: 

\start{Size of comparable corpora}
We sample the same percentage of comparable examples from AVATAR-comp and Gen-comp and combine the sampled data.
\ours is finetuned with at most 5 references in all trials.
As shown in \autoref{fig:hyper_cc}, as we increase the size of comparable corpora, \ours (\CC Only) always has better performance.
While the trends are not monotonic for \ours, it still obtains the best performance with full comparable corpora.

\start{Number of references}
We also examine the effect of the maximum number of references for each parallel example in finetuning.
We observe that the performance of \ours does not always increase when finetuned with more references.
The reason might be our model may generate more unique correct translations for some training examples than others.
As a result, the training signals the model received from different examples could be unbalanced, especially when the maximum number of references for each example is large.

\subsection{Case Studies}
We show a constructed comparable example in \autoref{fig:case}. More case studies for comparable corpora and generated references can be found in Appendices~\ref{sec:app_case_cc} and \ref{sec:app_case_multi}.
As shown in \autoref{fig:case}, both the Java function extracted from GitHub and the generated Python function remove a config by name and return this config.
The only difference is the Java function obtains the name of each config by a helper function, while the generated Python function directly accesses the ``\texttt{name}'' attribute.
We also observe that this function is likely to belong to a large software project, which has a different nature from coding problems (e.g., the example in \autoref{example_intro}).
As a result, combining collected and generated comparable examples provides diverse training signals to our model.
\section{Conclusion and Future Work}
We present two data augmentation techniques for code translation: comparable corpora and multiple references.
We study multiple ways of building comparable corpora, such as feeding natural language code documentation to code generation models.
Additionally, we generate multiple reference translations using our model and filter references using automatically generated unit tests.
Experiments on translation between 6 language pairs show that our two techniques improve CodeT5 substantially, by an average of 7.5\% CA@1.
Our analyses further show that after training on comparable corpora, the model learns to generate more fluent code with the same types of constructs. With multiple references, the model learns to generate a larger number of unique correct translations.

\edit{Following our analyses of what the model learns from comparable corpora, future work may conduct a more comprehensive study on what the model learns in code translation.
One may also explore combining our data augmentation techniques with data distillation from large language models (LLMs), where LLM may generate data with higher quality, but our techniques are less expensive.}

\section*{Limitations}
Despite the empirical advantages of using comparable corpora (\CC) shown in our work, there are some inherent biases and limitations in how we collect and construct them.
The collected \CC are from competitive programming websites, leading to a biased data distribution.
The \CC constructed using code generation models are also biased by the training data seen by these models and can potentially contain errors.
Furthermore, we only evaluate methods on the TransCoder dataset, which is currently the largest code translation dataset with test cases.
The TransCoder dataset only contains standalone functions that don't contain any imports outside the standard libraries for each language.
Translation of longer code with arbitrary external modules is an extension we plan to explore in future work. 
Another possible risk in our system is that the data may also contain information that uniquely identifies users or is offensive. For example, we generate comparable examples based on users' comments, which could contain inappropriate content.

\section*{Ethics Statement}
\start{License} We use public datasets AVATAR \cite{avatar}, xCodeEval \cite{xcodeeval}, and TransCoder \cite{transcoder} in our experiments. The data of AVATAR, xCodeEval, and TransCoder are all distributed under a CC BY-NC 4.0 license.

\start{Carbon Footprint} We avoided the usage of large language models when constructing our models. 
\ours has the same architecture as CodeT5-base, which has 220 million parameters.
The two models we use to construct comparable corpora have the same architecture as CodeT5-base and CodeT5-large, which have 220 million and 770 million parameters, respectively.
We train \ours first on the comparable corpora and then the parallel data.
Training on comparable corpora took 4-5 hours on average and training on parallel data took less than one hour on one \texttt{RTX A6000} GPU.
Therefore, training \ours results in approximately 0.78kg of carbon emission into the environment.\footnote{Estimations were conducted using the MachineLearning Impact calculator presented in \citet{CO2computation}. We use Amazon Web Services as the provider.}

\bibliography{anthology,custom}

\begin{thebibliography}{33}
\expandafter\ifx\csname natexlab\endcsname\relax\def\natexlab#1{#1}\fi

\bibitem[{Ahmad et~al.(2021{\natexlab{a}})Ahmad, Chakraborty, Ray, and
  Chang}]{plbart}
Wasi Ahmad, Saikat Chakraborty, Baishakhi Ray, and Kai-Wei Chang.
  2021{\natexlab{a}}.
\newblock \href {https://www.aclweb.org/anthology/2021.naacl-main.211} {Unified
  pre-training for program understanding and generation}.
\newblock In \emph{Proceedings of the 2021 Conference of the North American
  Chapter of the Association for Computational Linguistics: Human Language
  Technologies}, pages 2655--2668, Online. Association for Computational
  Linguistics.

\bibitem[{Ahmad et~al.(2022)Ahmad, Chakraborty, Ray, and Chang}]{SandG}
Wasi~Uddin Ahmad, Saikat Chakraborty, Baishakhi Ray, and Kai-Wei Chang. 2022.
\newblock Summarize and generate to back-translate: Unsupervised translation of
  programming languages.
\newblock In \emph{Conference of the European Chapter of the Association for
  Computational Linguistics}.

\bibitem[{Ahmad et~al.(2021{\natexlab{b}})Ahmad, Tushar, Chakraborty, and
  Chang}]{avatar}
Wasi~Uddin Ahmad, Md~Golam~Rahman Tushar, Saikat Chakraborty, and Kai-Wei
  Chang. 2021{\natexlab{b}}.
\newblock Avatar: A parallel corpus for java-python program translation.
\newblock \emph{arXiv preprint arXiv:2108.11590}.

\bibitem[{Austin et~al.(2021)Austin, Odena, Nye, Bosma, Michalewski, Dohan,
  Jiang, Cai, Terry, Le, and Sutton}]{MBPP}
Jacob Austin, Augustus Odena, Maxwell Nye, Maarten Bosma, Henryk Michalewski,
  David Dohan, Ellen Jiang, Carrie Cai, Michael Terry, Quoc Le, and Charles
  Sutton. 2021.
\newblock \href {http://arxiv.org/abs/2108.07732} {Program synthesis with large
  language models}.

\bibitem[{Cassano et~al.(2022)Cassano, Gouwar, Nguyen, Nguyen, Phipps-Costin,
  Pinckney, Yee, Zi, Anderson, Feldman, Guha, Greenberg, and Jangda}]{multiple}
Federico Cassano, John Gouwar, Daniel Nguyen, Sydney Nguyen, Luna
  Phipps-Costin, Donald Pinckney, Ming-Ho Yee, Yangtian Zi, Carolyn~Jane
  Anderson, Molly~Q Feldman, Arjun Guha, Michael Greenberg, and Abhinav Jangda.
  2022.
\newblock \href {http://arxiv.org/abs/2208.08227} {Multipl-e: A scalable and
  extensible approach to benchmarking neural code generation}.

\bibitem[{Chen et~al.(2018)Chen, Liu, and Song}]{TreetotreeNN}
Xinyun Chen, Chang Liu, and Dawn~Xiaodong Song. 2018.
\newblock Tree-to-tree neural networks for program translation.
\newblock In \emph{Neural Information Processing Systems}.

\bibitem[{Etchegoyhen and Gete(2020)}]{handle_with_care_comp_corpora}
T.~Etchegoyhen and Harritxu Gete. 2020.
\newblock Handle with care: A case study in comparable corpora exploitation for
  neural machine translation.
\newblock In \emph{International Conference on Language Resources and
  Evaluation}.

\bibitem[{Feng et~al.(2020)Feng, Guo, Tang, Duan, Feng, Gong, Shou, Qin, Liu,
  Jiang, and Zhou}]{codebert}
Zhangyin Feng, Daya Guo, Duyu Tang, Nan Duan, Xiaocheng Feng, Ming Gong, Linjun
  Shou, Bing Qin, Ting Liu, Daxin Jiang, and Ming Zhou. 2020.
\newblock \href {https://doi.org/10.18653/v1/2020.findings-emnlp.139}
  {{C}ode{BERT}: A pre-trained model for programming and natural languages}.
\newblock In \emph{Findings of the Association for Computational Linguistics:
  EMNLP 2020}, pages 1536--1547, Online. Association for Computational
  Linguistics.

\bibitem[{Gete and Etchegoyhen(2022)}]{NMTComparableCorpora}
Harritxu Gete and Thierry Etchegoyhen. 2022.
\newblock \href {https://doi.org/10.1007/s10579-021-09572-2} {Making the most
  of comparable corpora in neural machine translation: a case study}.
\newblock \emph{Language Resources and Evaluation}, 56(3):943--971.

\bibitem[{Husain et~al.(2020)Husain, Wu, Gazit, Allamanis, and
  Brockschmidt}]{codesearchnet}
Hamel Husain, Ho-Hsiang Wu, Tiferet Gazit, Miltiadis Allamanis, and Marc
  Brockschmidt. 2020.
\newblock \href {http://arxiv.org/abs/1909.09436} {Codesearchnet challenge:
  Evaluating the state of semantic code search}.

\bibitem[{Irvine(2014)}]{low_resource_comp_corpora}
Ann Irvine. 2014.
\newblock Using comparable corpora to augment statistical machine translation
  models in low resource settings.

\bibitem[{Iyer et~al.(2018)Iyer, Konstas, Cheung, and Zettlemoyer}]{concode}
Srinivasan Iyer, Ioannis Konstas, Alvin Cheung, and Luke Zettlemoyer. 2018.
\newblock \href {https://www.aclweb.org/anthology/D18-1192} {Mapping language
  to code in programmatic context}.
\newblock In \emph{Proceedings of the 2018 Conference on Empirical Methods in
  Natural Language Processing}. Association for Computational Linguistics.

\bibitem[{Karaivanov et~al.(2014)Karaivanov, Raychev, and
  Vechev}]{Karaivanov2014PhraseBasedST}
Svetoslav Karaivanov, Veselin Raychev, and Martin~T. Vechev. 2014.
\newblock Phrase-based statistical translation of programming languages.
\newblock \emph{Proceedings of the 2014 ACM International Symposium on New
  Ideas, New Paradigms, and Reflections on Programming \& Software}.

\bibitem[{Khan et~al.(2023)Khan, Bari, Do, Wang, Parvez, and Joty}]{xcodeeval}
Mohammad Abdullah~Matin Khan, M~Saiful Bari, Xuan~Long Do, Weishi Wang,
  Md~Rizwan Parvez, and Shafiq Joty. 2023.
\newblock \href {http://arxiv.org/abs/2303.03004} {xcodeeval: A large scale
  multilingual multitask benchmark for code understanding, generation,
  translation and retrieval}.

\bibitem[{Khayrallah et~al.(2020)Khayrallah, Thompson, Post, and
  Koehn}]{simulated_multi_ref}
Huda Khayrallah, Brian Thompson, Matt Post, and Philipp Koehn. 2020.
\newblock Simulated multiple reference training improves low-resource machine
  translation.
\newblock In \emph{Conference on Empirical Methods in Natural Language
  Processing}.

\bibitem[{Lachaux et~al.(2021)Lachaux, Roziere, Szafraniec, and Lample}]{dobf}
Marieanne Lachaux, Baptiste Roziere, Marc Szafraniec, and Guillaume Lample.
  2021.
\newblock {DOBF}: A deobfuscation pre-training objective for programming
  languages.
\newblock In \emph{Advances in Neural Information Processing Systems}.

\bibitem[{Lacoste et~al.(2019)Lacoste, Luccioni, Schmidt, and
  Dandres}]{CO2computation}
Alexandre Lacoste, Alexandra Luccioni, Victor Schmidt, and Thomas Dandres.
  2019.
\newblock \href {http://arxiv.org/abs/1910.09700} {Quantifying the carbon
  emissions of machine learning}.

\bibitem[{Le et~al.(2022)Le, Wang, Gotmare, Savarese, and Hoi}]{coderl}
Hung Le, Yue Wang, Akhilesh~Deepak Gotmare, Silvio Savarese, and Steven Hoi.
  2022.
\newblock \href {https://openreview.net/forum?id=WaGvb7OzySA} {Code{RL}:
  Mastering code generation through pretrained models and deep reinforcement
  learning}.
\newblock In \emph{Advances in Neural Information Processing Systems}.

\bibitem[{Lu et~al.(2021)Lu, Guo, Ren, Huang, Svyatkovskiy, Blanco, Clement,
  Drain, Jiang, Tang, Li, Zhou, Shou, Zhou, Tufano, GONG, Zhou, Duan,
  Sundaresan, Deng, Fu, and LIU}]{codexglue}
Shuai Lu, Daya Guo, Shuo Ren, Junjie Huang, Alexey Svyatkovskiy, Ambrosio
  Blanco, Colin Clement, Dawn Drain, Daxin Jiang, Duyu Tang, Ge~Li, Lidong
  Zhou, Linjun Shou, Long Zhou, Michele Tufano, MING GONG, Ming Zhou, Nan Duan,
  Neel Sundaresan, Shao~Kun Deng, Shengyu Fu, and Shujie LIU. 2021.
\newblock Code{XGLUE}: A machine learning benchmark dataset for code
  understanding and generation.
\newblock In \emph{Thirty-fifth Conference on Neural Information Processing
  Systems Datasets and Benchmarks Track (Round 1)}.

\bibitem[{Munteanu and Marcu(2005)}]{first_comp_corpora}
Dragos~Stefan Munteanu and Daniel Marcu. 2005.
\newblock Improving machine translation performance by exploiting non-parallel
  corpora.
\newblock \emph{Computational Linguistics}.

\bibitem[{Nguyen et~al.(2013)Nguyen, Nguyen, and Nguyen}]{Nguyen2013LexicalSM}
Anh~Tuan Nguyen, Tung~Thanh Nguyen, and Tien~Nhut Nguyen. 2013.
\newblock Lexical statistical machine translation for language migration.
\newblock In \emph{ESEC/FSE 2013}.

\bibitem[{Nguyen et~al.(2015)Nguyen, Nguyen, and
  Nguyen}]{Nguyen2015DivideandConquerAF}
Anh~Tuan Nguyen, Tung~Thanh Nguyen, and Tien~Nhut Nguyen. 2015.
\newblock Divide-and-conquer approach for multi-phase statistical migration for
  source code (t).
\newblock \emph{2015 30th IEEE/ACM International Conference on Automated
  Software Engineering (ASE)}.

\bibitem[{Oda et~al.(2015)Oda, Fudaba, Neubig, Hata, Sakti, Toda, and
  Nakamura}]{Oda2015LearningTG}
Yusuke Oda, Hiroyuki Fudaba, Graham Neubig, Hideaki Hata, Sakriani Sakti,
  Tomoki Toda, and Satoshi Nakamura. 2015.
\newblock Learning to generate pseudo-code from source code using statistical
  machine translation (t).
\newblock \emph{2015 30th IEEE/ACM International Conference on Automated
  Software Engineering (ASE)}.

\bibitem[{Papineni et~al.(2002)Papineni, Roukos, Ward, and Zhu}]{BLEU}
Kishore Papineni, Salim Roukos, Todd Ward, and Wei-Jing Zhu. 2002.
\newblock \href {https://doi.org/10.3115/1073083.1073135} {Bleu: A method for
  automatic evaluation of machine translation}.
\newblock In \emph{Proceedings of the 40th Annual Meeting on Association for
  Computational Linguistics}, ACL '02. Association for Computational
  Linguistics.

\bibitem[{Paramita et~al.(2013)Paramita, Guthrie, Kanoulas, Gaizauskas, Clough,
  and Sanderson}]{methods_collect_comp_corpora}
Monica~Lestari Paramita, David Guthrie, E.~Kanoulas, Robert~J. Gaizauskas,
  Paul~D. Clough, and Mark Sanderson. 2013.
\newblock Methods for collection and evaluation of comparable documents.
\newblock In \emph{Building and Using Comparable Corpora}.

\bibitem[{Phan et~al.(2017)Phan, Nguyen, Nguyen, and
  Nguyen}]{Phan2017StatisticalMO}
Hung~Dang Phan, Anh~Tuan Nguyen, Trong~Duc Nguyen, and Tien~Nhut Nguyen. 2017.
\newblock Statistical migration of api usages.
\newblock \emph{2017 IEEE/ACM 39th International Conference on Software
  Engineering Companion (ICSE-C)}, pages 47--50.

\bibitem[{Puri et~al.(2021)Puri, Kung, Janssen, Zhang, Domeniconi, Zolotov,
  Dolby, Chen, Choudhury, Decker, Thost, Buratti, Pujar, Ramji, Finkler,
  Malaika, and Reiss}]{codenet}
Ruchir Puri, David~S. Kung, Geert Janssen, Wei Zhang, Giacomo Domeniconi,
  Vladimir Zolotov, Julian Dolby, Jie Chen, Mihir Choudhury, Lindsey Decker,
  Veronika Thost, Luca Buratti, Saurabh Pujar, Shyam Ramji, Ulrich Finkler,
  Susan Malaika, and Frederick Reiss. 2021.
\newblock \href {http://arxiv.org/abs/2105.12655} {Codenet: A large-scale ai
  for code dataset for learning a diversity of coding tasks}.

\bibitem[{Qin and Specia(2015)}]{truly_exploring_multi_ref}
Ying Qin and Lucia Specia. 2015.
\newblock Truly exploring multiple references for machine translation
  evaluation.
\newblock In \emph{Proceedings of the 18th Annual Conference of the {E}uropean
  Association for Machine Translation}.

\bibitem[{Ren et~al.(2020)Ren, Guo, Lu, Zhou, Liu, Tang, Sundaresan, Zhou,
  Blanco, and Ma}]{codebleu}
Shuo Ren, Daya Guo, Shuai Lu, Long Zhou, Shujie Liu, Duyu Tang, Neel
  Sundaresan, Ming Zhou, Ambrosio Blanco, and Shuai Ma. 2020.
\newblock \href {http://arxiv.org/abs/2009.10297} {Codebleu: a method for
  automatic evaluation of code synthesis}.

\bibitem[{Roziere et~al.(2020)Roziere, Lachaux, Chanussot, and
  Lample}]{transcoder}
Baptiste Roziere, Marie-Anne Lachaux, Lowik Chanussot, and Guillaume Lample.
  2020.
\newblock Unsupervised translation of programming languages.
\newblock \emph{Advances in Neural Information Processing Systems}, 33.

\bibitem[{Roziere et~al.(2022)Roziere, Zhang, Charton, Harman, Synnaeve, and
  Lample}]{transcoderST}
Baptiste Roziere, Jie~M. Zhang, Francois Charton, Mark Harman, Gabriel
  Synnaeve, and Guillaume Lample. 2022.
\newblock \href {http://arxiv.org/abs/2110.06773} {Leveraging automated unit
  tests for unsupervised code translation}.

\bibitem[{Wang et~al.(2021)Wang, Wang, Joty, and Hoi}]{codet5}
Yue Wang, Weishi Wang, Shafiq Joty, and Steven~C.H. Hoi. 2021.
\newblock {C}ode{T}5: Identifier-aware unified pre-trained encoder-decoder
  models for code understanding and generation.
\newblock In \emph{Proceedings of the 2021 Conference on Empirical Methods in
  Natural Language Processing}.

\bibitem[{Wołk et~al.(2015)Wołk, Rejmund, and
  Marasek}]{harvesting_comp_corpora}
Krzysztof Wołk, Emilia Rejmund, and Krzysztof Marasek. 2015.
\newblock Harvesting comparable corpora and mining them for equivalent
  bilingual sentences using statistical classification and analogy-based
  heuristics.
\newblock \emph{ArXiv}, abs/1511.06285.

\end{thebibliography}
\bibliographystyle{acl_natbib}

\newpage
\appendix

\section{Appendix}
\label{sec:appendix}

\subsection{Experimental Details}
\label{sec:implementation}

\start{Implementation details}
To generate the comparable corpora dataset (denoted as ``Gen-comp'' in \autoref{tab:dataset}), for Java $\leftrightarrow$ Python translation, we obtain Java functions with Docstrings from CodeSearchNet \cite{codesearchnet} and use CodeRL \cite{coderl} finetuned on MBPP \cite{MBPP} to generate Python programs.
For Cpp $\leftrightarrow$ Java translation, we obtain Cpp solutions with problem descriptions from CodeNet \cite{codenet} and use CodeT5 \cite{codet5} finetuned on CONCODE \cite{concode} to generate Java programs.
For Cpp $\leftrightarrow$ Python translation, we again obtain Cpp programs from CodeNet and use the finetuned CodeRL model to generate Python programs.

To train \ours, we tune the number of source/target programs per problem in the range of [1, 3, 5] and tune the maximum number of generated references for finetuning in the range of [5, 10]. 
For the training on both comparable corpora and AVATAR-para, we train \ours with a learning rate of 1e-4 and batch size of 32 for at most 20 epochs.
As for the baselines, we finetune CodeT5 with a learning rate of 1e-4 and batch size of 32 for at most 20 epochs.
We finetune TransCoder-ST-ft with a learning rate of 1e-4 and batch size of 64 for at most 20 epochs.

\start{Baselines}
We compare \ours with unsupervised and self-supervised models, including TransCoder \cite{transcoder}, DOBF \cite{dobf}, and TransCoder-ST \cite{transcoderST}. 
We also compare with supervised models, which are initialized from CodeBERT \cite{codebert}, PLBART \cite{plbart}, CodeT5 \cite{codet5}, or TransCoder-ST \cite{transcoderST} (denoted as TransCoder-ST-ft) and finetuned on AVATAR-para.
We only compare with DOBF, CodeBERT, and PLBART on Java $\leftrightarrow$ Python translation because these models are not pretrained on Cpp.

\subsection{Statistical Significance Test}
\label{sec:ttest}
We present the t-test results in \autoref{tab:t-test}, where we run each experiment with 3 random seeds when finetuning on AVATAR-para.

\begin{table*}[]
\centering
\resizebox{0.9\textwidth}{!}{
\begin{tabular}{lcccccc}
\toprule
 & \multicolumn{3}{c}{\bf Java-to-Python} 
 & \multicolumn{3}{c}{\bf Python-to-Java} \\
\cmidrule(lr){2-4} \cmidrule(lr){5-7}
 & \bf BLEU & \bf CB & \bf CA@1 & \bf BLEU & \bf CB & \bf CA@1 \\
 \midrule
\bf Best baseline & 79.9 & 73.2 & 69.4 & 81.4 & 78.4 & 62.0 \\
\bf \ours & 82.1 $\pm$ 0.4** & 76.1 $\pm$ 1.4** & 72.1 $\pm$ 1.6** & 84.3 $\pm$ 0.4** & 82.0 $\pm$ 0.5** & 64.6 $\pm$ 1.0** \\
\bottomrule
\toprule
 & \multicolumn{3}{c}{\bf C++-to-Python} 
 & \multicolumn{3}{c}{\bf Python-to-C++} \\
\cmidrule(lr){2-4} \cmidrule(lr){5-7}
 & \bf BLEU & \bf CB & \bf CA@1 & \bf BLEU & \bf CB & \bf CA@1 \\
 \midrule
\bf Best baseline & 82.9 & 75.4 & 62.5 & 79.8 & 77.9 & 59.1 \\
\bf \ours & 83.8 $\pm$ 0.6** & 76.8 $\pm$ 0.3** & 64.6 $\pm$ 0.8** & 82.1 $\pm$ 0.9** & 79.8 $\pm$ 0.5** & 63.2 $\pm$ 1.4** \\
\bottomrule
\toprule
 & \multicolumn{3}{c}{\bf Java-to-C++} 
 & \multicolumn{3}{c}{\bf C++-to-Java} \\
\cmidrule(lr){2-4} \cmidrule(lr){5-7}
 & \bf BLEU & \bf CB & \bf CA@1 & \bf BLEU & \bf CB & \bf CA@1 \\
 \midrule
\bf Best baseline & 89.3 & 88.5 & 84.6 & 90.9 & 90.1 & 68.3 \\
\bf \ours & 89.9 $\pm$ 0.4* & 88.1 $\pm$ 0.2 & 83.7 $\pm$ 1.1 
& 91.1 $\pm$ 0.4 & 90.3 $\pm$ 0.1** & 69.9 $\pm$ 1.1* \\
\bottomrule
\end{tabular}
}
\caption{T-test results. We copy the results from the papers of the best baselines and run \ours for 3 different random seeds. **means significant results with p-value < 0.01. *means significant results with p-value < 0.05.}
\label{tab:t-test}
\end{table*}

\subsection{Case Studies for Naturally Existing and Generated Comparable Corpora}
\label{sec:app_case_cc}
\begin{figure*}[htb]
    \centering
    \includegraphics[width=\linewidth]{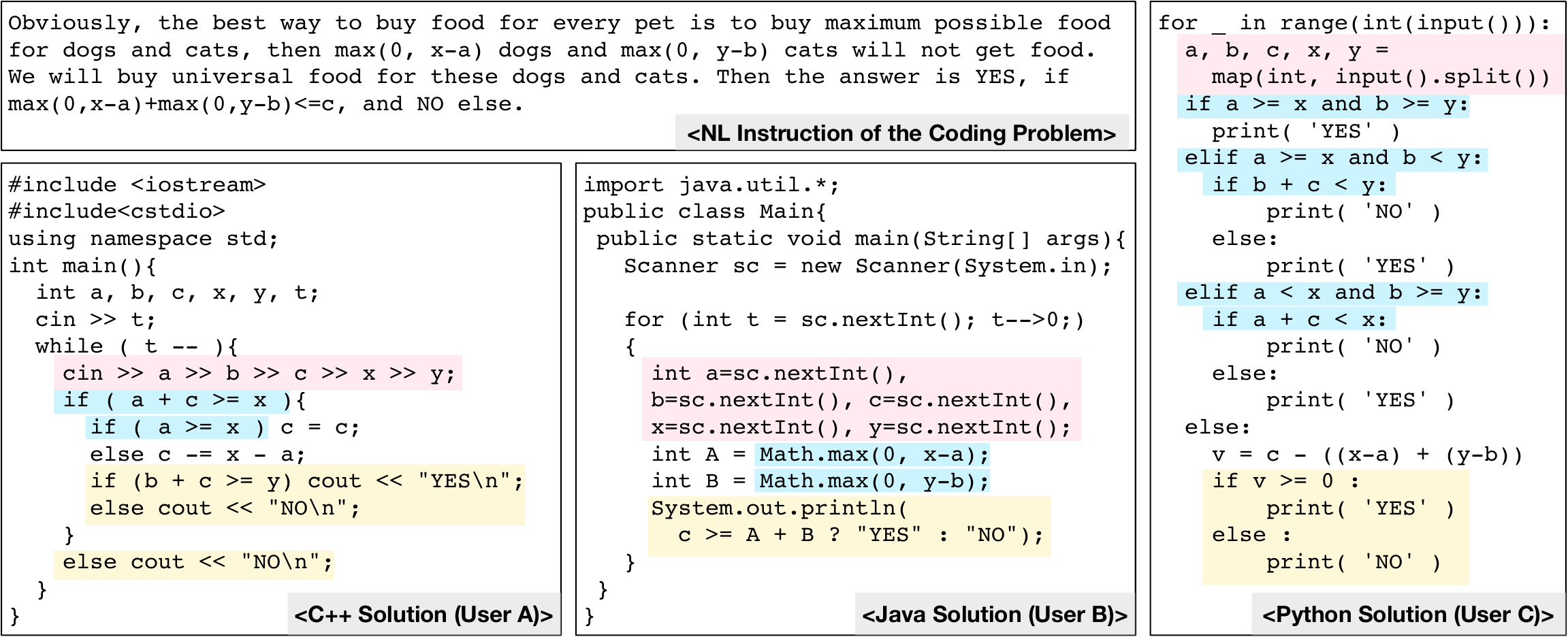}
    \caption{A comparable example collected from a coding problem. We highlight the lines in the user-written C++, Java, and Python programs that can be matched.}
    \label{fig:app_collect}
\end{figure*}

\begin{figure*}[htb]
    \centering
    \includegraphics[width=\linewidth]{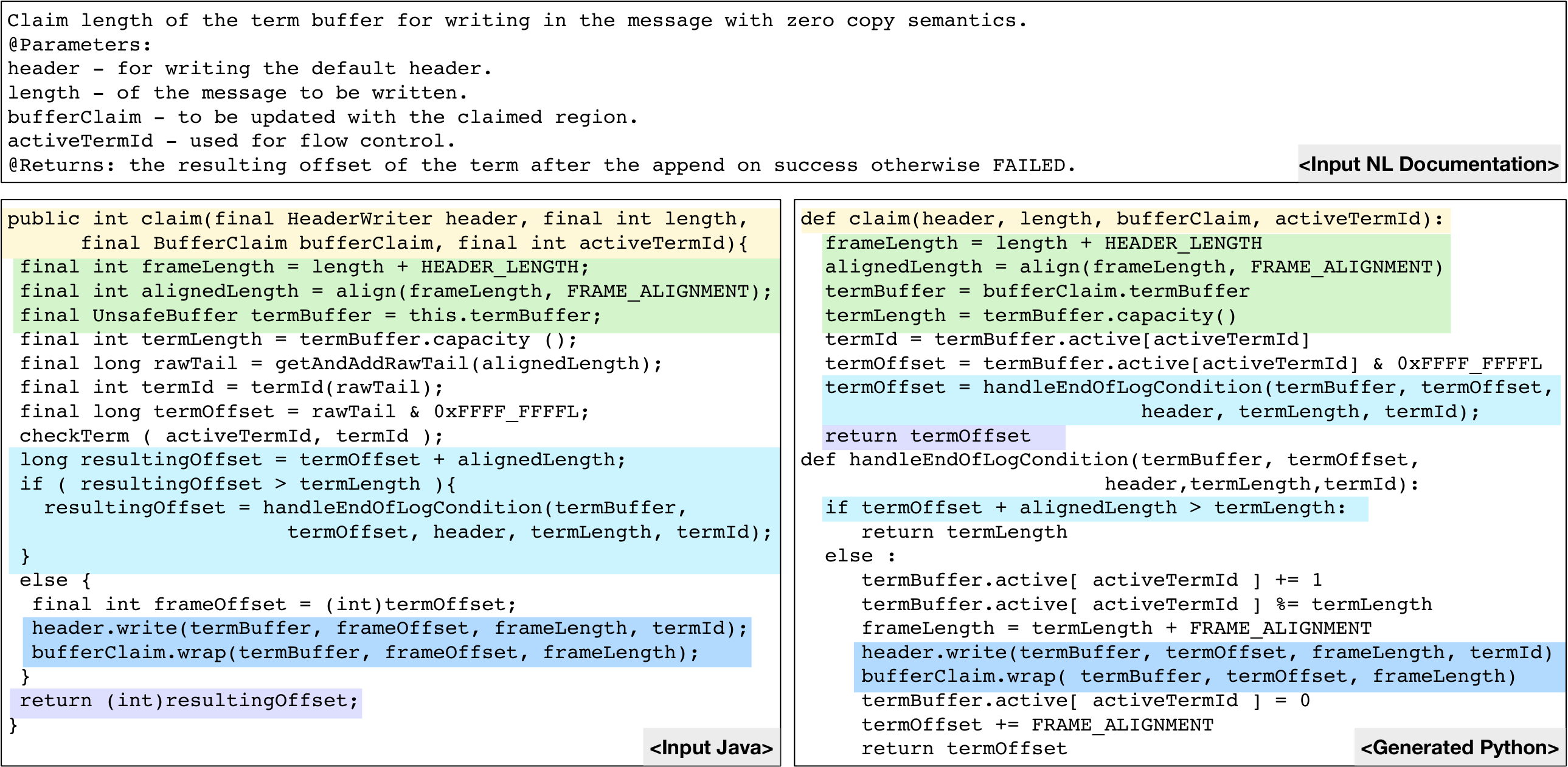}
    \caption{A comparable example generated by our method. The natural language documentation is the docstring of the Java program. We highlight the lines in the Java program and the generated Python program that can be matched.}
    \label{fig:app_generate}
\end{figure*}

We show case studies of naturally existing and generated comparable examples in \autoref{fig:app_collect} and \autoref{fig:app_generate}.
The example in \autoref{fig:app_collect} is from the xCodeEval dataset, which contains different users' submissions of the ``Food for Animals'' problem on Codeforces.
In \autoref{fig:app_generate}, the Java function and its docstring are from the CodeSearchNet dataset. The Python program is generated by our method.

\start{Quality of the Generated Comparable Example}
As shown in \autoref{fig:app_generate}, the Python program we generate has similar functionality as the input Java program.
Specifically, both programs define a list of variables, compare the sum of term offset and aligned length with the term length, and return an offset of the term.
The major differences are that the Java program calls a global function ``\texttt{getAndAddRawTail}'' to compute the value of ``\texttt{rawTail}'', while the generated Python program calls a class function ``\texttt{termBuffer.active()}''. Both functions are not defined in the context.
Also, the Java program calls another function, ``\texttt{handleEndOfLogCondition}'', without defining it, while our method also generates the content of ``\texttt{handleEndOfLogCondition}''. 

We notice that a large percentage of differences between the generated and input program are due to calling functions without presenting their definitions in the input.
Such input programs contain limited information on the purpose of these functions. As a result, it is challenging for the code generation model to generate code with exactly the same functionality.

\start{Comparison Between Naturally Existing and Generated Comparable Examples}
We can observe that the naturally existing and generated comparable examples are different in several ways. For instance, the naturally existing comparable examples are mainly collected from coding problems, while the generated examples can belong to other sources, such as a large software project.
In addition, the programs in the naturally existing example are self-contained, while both the Java and Python programs in the generated example contain user-defined classes and external functions that are defined elsewhere, including ``\texttt{HeaderWriter}'', ``\texttt{HeaderWriter.write()}'', and ``\texttt{align()}''.
Besides, programs in one naturally existing comparable example typically have the same output format (e.g., ``\texttt{YES}'' or ``\texttt{NO}'' in this case), while programs we generate from the documentation may have different output formats, especially when the docstring is unclear about the return value.
With all the differences, the combination of naturally existing and generated comparable corpora cover programs with a large variety of styles and domains, which provides diverse training signals to our model.

\subsection{Case Studies for Model-generated References}
\label{sec:app_case_multi}
\begin{figure*}[t]
    \centering
    \includegraphics[width=\linewidth]{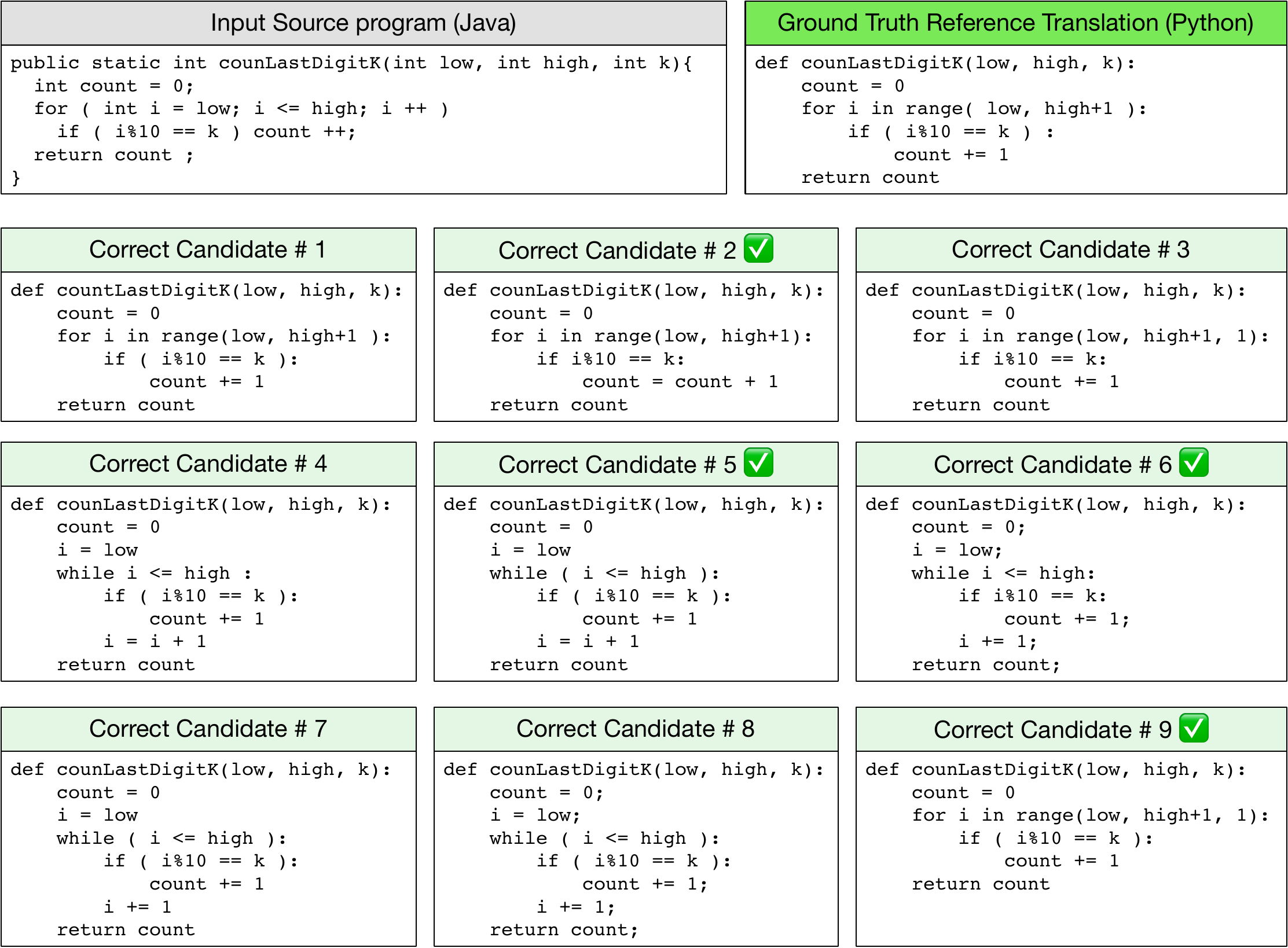}
    \caption{An example of the multiple references we generate. We show the top 9 correct translations generated by our model with the highest probabilities. Candidates with a checkmark are those we select to add to the reference set. We add these translations one by one based on their string edit distance with all the existing translations.}
    \label{fig:app_multi}
\end{figure*}

\autoref{fig:app_multi} shows an example of the multiple references generated by \ours. The example is from the AVATAR-para dataset. We use our model to generate 50 candidate translations, 23 of which are correct (i.e., have exactly the same output as the source program on all the automatically generated test cases). We show the 9 correct translations with the highest probabilities, 4 of which are selected to the reference set.

\start{Quality of the Generated References}
In this example, the source program contains a for loop and an if statement. Note that the for loop has ``$\leq$ \texttt{high}'' as the end condition. When we translate it to the ``\texttt{for \$ITER\_VAR in range()}'' grammar in Python, we have to instead use ``\texttt{high + 1}'' as the end value. This is correctly handled in several different candidate translations.

\start{Diversity of the Generated References}
There are two main options for translating a loop: using either a for loop or a while loop. 
In the example in \autoref{fig:app_multi}, the selected reference set contains 2 translations with a while loop and 3 with a for loop, including the ground truth translation.

We can observe that the differences between two candidates can be as small as adding or deleting parentheses and semicolons.
For instance, the only difference between Candidate \#3 and Candidate \#9 is the parentheses around ``\texttt{i\%10 == k}''.
In this example, by computing the string edit distance, we avoid having both Candidate \#3 and \#9 in the reference set.

\end{document}